\documentclass{article}

\PassOptionsToPackage{numbers, compress}{natbib}

    \usepackage[preprint]{neurips_2024}

\usepackage[utf8]{inputenc} 
\usepackage[T1]{fontenc}    
\usepackage{hyperref}       
\usepackage{url}            
\usepackage{booktabs}       
\usepackage{amsfonts}       
\usepackage{nicefrac}       
\usepackage{microtype}      
\usepackage{xcolor,colortbl}         
\usepackage{graphicx}
\usepackage{multirow}
\usepackage{amssymb}
\usepackage{pifont}
\usepackage{wrapfig}
\newcommand{\hlrow}{\rowcolor{black!6}}
\usepackage{xcolor,colortbl}
\usepackage{multirow}
\usepackage{subcaption}
\usepackage{tablefootnote}
\usepackage{algorithm}
\usepackage{algpseudocode}
\usepackage{amsmath}

\definecolor{yzybest}{rgb}{0.96, 0.57, 0.58}
\definecolor{yzysecond}{rgb}{0.98, 0.78, 0.57}
\definecolor{yzythird}{rgb}{1.0, 1.0, 0.56}
\definecolor{myblue}{HTML}{1f77b4}
\definecolor{myorange2}{HTML}{ff7f03}
\definecolor{myred}{HTML}{FF0100}
\definecolor{gscolor}{rgb}{1.0,0.6,0.0} %
\definecolor{todo}{rgb}{1.0, 0.0, 0.0}
\definecolor{mygreen}{RGB}{34, 139, 34}
\definecolor{mygray}{gray}{0.9}

\newcommand{\oursplit}{ABE-Split}
\newcommand{\decrease}[2]{\shortstack{#1 \\ \tiny \textcolor{red}{(#2)}}}
\newcommand{\increase}[2]{\shortstack{#1 \\ \tiny \textcolor{mygreen}{(#2)}}}
\newcommand{\stay}[1]{\shortstack{#1 \\ \tiny \textcolor{gray}{(\ \ \ \ -- \ \ \ )}}}
\newcommand{\rain}{{\includegraphics[scale=0.04]
{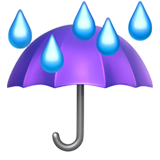}}}

\title{Relaxing Accurate Initialization Constraint for \\ 3D Gaussian Splatting}

\author{
\textbf{Jaewoo Jung}$^\rain$, 
~\quad
\textbf{Jisang Han}$^\rain$,
~\quad
\textbf{Honggyu An}$^\rain$,
\\
\textbf{Jiwon Kang}$^\rain$,
~\quad
\textbf{Seonghoon Park}$^\rain$,
~\quad
\textbf{Seungryong Kim}
\\\\
Korea University \\
\textcolor{blue}{\url{https://ku-cvlab.github.io/RAIN-GS}}
}

\begin{document}

\maketitle

\begingroup
\renewcommand{\thefootnote}{}
\footnotetext{$\rain$: Equal contribution.}
\endgroup

\begin{figure*}[h]
\begin{center}
\vspace{-30pt}
\includegraphics[width=1\linewidth]{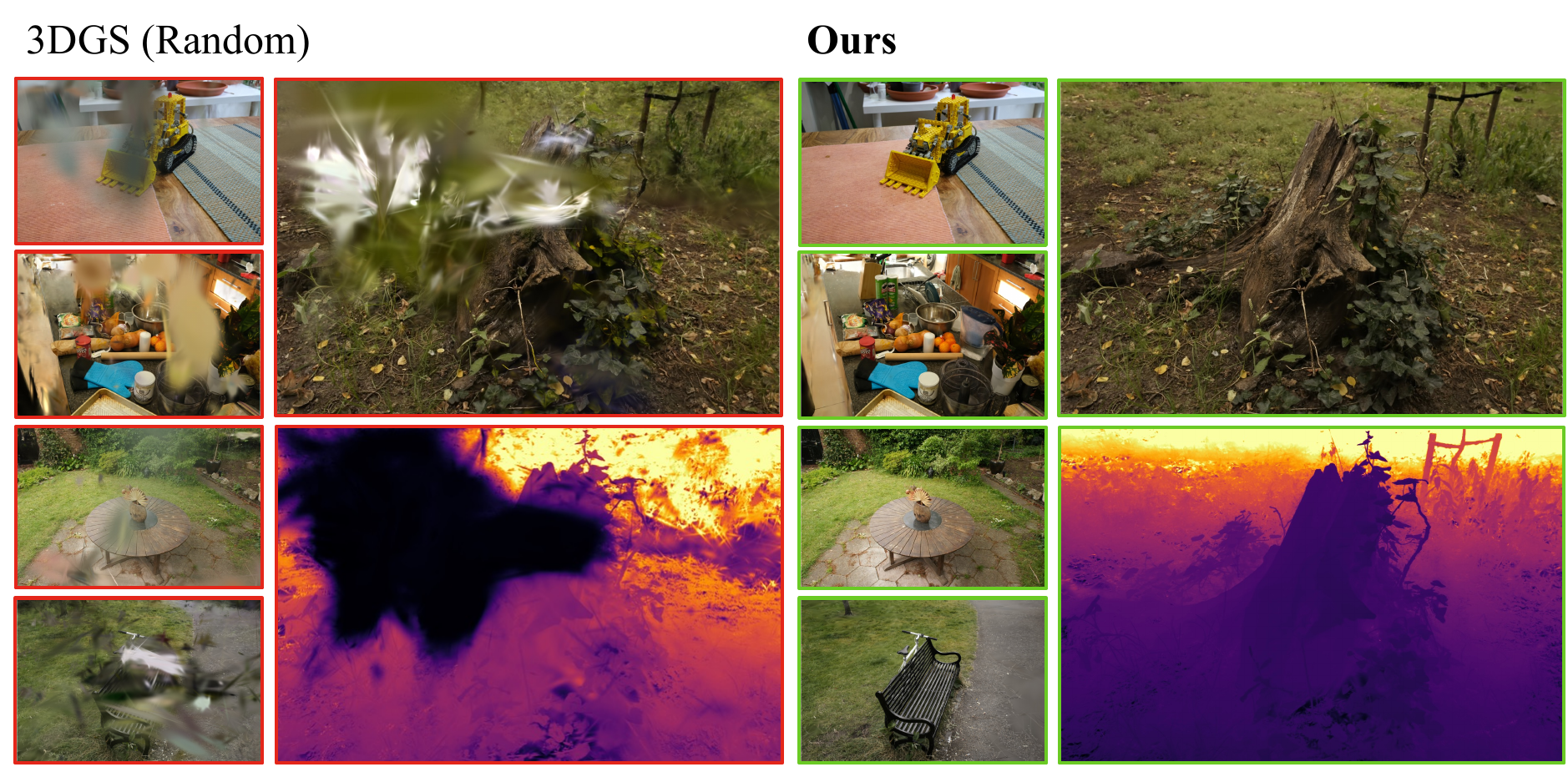}
\end{center}
    \caption{\textbf{Effectiveness of our simple strategy.} \textcolor{red}{Left} and \textcolor{green}{right} show the results from 3DGS~\cite{kerbl20233d} and ours trained with randomly initialized point cloud respectively. Transition from 3DGS to ours simply requires our strategy consisted of sparse-large-variance (SLV) random initialization, progressive Gaussian low-pass filtering, and adaptive bound-expanding split (ABE-Split) algorithm.}
    \label{fig:teaser}
\end{figure*}

\begin{abstract}
\vspace{-5pt}
3D Gaussian splatting (3DGS) has recently demonstrated impressive capabilities in real-time novel view synthesis and 3D reconstruction. However, 3DGS heavily depends on the accurate initialization derived from Structure-from-Motion (SfM) methods. When the quality of the initial point cloud deteriorates, such as in the presence of noise or when using randomly initialized point cloud, 3DGS often undergoes large performance drops. To address this limitation, we propose a novel optimization strategy dubbed \textbf{RAIN-GS} (\textbf{R}elaxing \textbf{A}ccurate \textbf{IN}itialization Constraint for 3D \textbf{G}aussian \textbf{S}platting). Our approach is based on an in-depth analysis of the original 3DGS optimization scheme and the analysis of the SfM initialization in the frequency domain. Leveraging simple modifications based on our analyses, \textbf{RAIN-GS} successfully trains 3D Gaussians from sub-optimal point cloud~(e.g., randomly initialized point cloud), effectively relaxing the need for accurate initialization. We demonstrate the efficacy of our strategy through quantitative and qualitative comparisons on multiple datasets, where \textbf{RAIN-GS} trained with random point cloud achieves performance on-par with or even better than 3DGS trained with accurate SfM point cloud.
\end{abstract}
\section{Introduction}
\label{sec:intro}

Novel view synthesis is one of the essential tasks in computer vision and computer graphics, aiming to render novel views of a 3D scene given a set of images. It has a wide range of applications in various fields, including augmented reality and virtual reality~\cite{xu2023vr}, robotics~\cite{adamkiewicz2022vision}, and data generation~\cite{ge2022neural}. Neural radiance fields (NeRFs)~\cite{mildenhall2021nerf} have demonstrated remarkable success in this task, learning implicit representations that capture intricate 3D geometry and specular effects solely from images. However, NeRFs' reliance on multi-layer perceptrons (MLPs)~\cite{fridovich2023k, liu2020neural, reiser2021kilonerf,tancik2022blocknerf,zhang2020nerf} results in slow and computationally intensive volume rendering, hindering real-time applications.

Recently, 3D Gaussian splatting (3DGS)~\cite{kerbl20233d} has emerged as a compelling alternative for both high-quality results and real-time rendering. Unlike NeRFs training an MLP to model the scene with an implicit representation, 3DGS models the scene using explicit 3D Gaussians. The learned 3D Gaussians are then projected to the image space using an efficient CUDA-based differentiable tile rasterization~\cite{kopanas2021point, kerbl20233d} technique which enables the rendering of the Gaussians in real-time. 

Despite its remarkable results, 3DGS requires an additional input of initial point cloud compared to NeRFs. The quality of the initial point cloud is one of the essential requirements of 3DGS, showing large performance drops when trained with randomly initialized point cloud~\cite{kerbl20233d}. A similar type of performance degradation can occur in real-world scenarios, specifically in scenes where SfM techniques struggle to converge, such as scenes with symmetry, and textureless regions~\cite{bian2023nope,zhang2022relpose}. Also, accurate initial point cloud become a strict requirement even for situations where camera poses can be obtained through external sensors~\cite{geiger2013vision,sturm2012benchmark} or pre-defined such as in text-to-3D generation~\cite{tang2023dreamgaussian,yi2023gaussiandreamer}. In these cases, where the initial point cloud from SfM may contain noise or even unavailable, 3DGS requires an additional module to obtain the point cloud, adding extra computation cost~\cite{nichol2022point,wang2023dust3r}. 

In this work, we start with a natural question: \textit{``Why is accurate initial point cloud so important in 3D Gaussian splatting?''} and conduct an in-depth analysis of the original 3DGS optimization scheme and SfM initialization. Our analyses reveal that the original 3DGS optimization scheme struggles to \textit{transport} Gaussians further from their initialized locations, leading to performance degradation when starting from random or noisy SfM initialization. In addition, the analysis of the SfM initialization in the frequency domain reveals that SfM initialization provides the low-frequency components of the true distribution. From the provided low-frequency components, 3DGS clones and splits existing Gaussians, which is then used to learn the remaining high-frequency details of the scene~\cite{kerbl20233d}. In this process, the low-frequency components guide the Gaussians to successfully learn the target distribution in a \textit{coarse-to-fine} manner utilizing the clone/split process, without the need to transport the Gaussians to correct locations. Based on this observation, we propose that to train 3DGS with noisy or random point cloud, we need an effective strategy that can transport the Gaussians further from their initial positions and encourage the Gaussians to prioritize the learning of the coarse structure information, which will then guide the overall optimization process to learn in a similar coarse-to-fine manner.

Based on our analysis of accurate initial point cloud, we propose a novel optimization strategy called \textbf{RAIN-GS} (\textbf{R}elaxing \textbf{A}ccurate \textbf{IN}itialization Constraint for 3D \textbf{G}aussian \textbf{S}platting), which is composed of three key components: \textbf{1)} a novel initialization method starting with sparse Gaussians with large variance, \textbf{2)} progressive Gaussian low-pass filtering utilized in the rendering process, and \textbf{3)} a novel adaptive bound-expanding split (\oursplit) algorithm utilized in the adaptive density control. We demonstrate that RAIN-GS effectively relaxes the requirement of accurate initialization for 3DGS, as applying RAIN-GS to random point cloud successfully priorities the learning of low-frequency components and guides the Gaussians to learn in a \textit{coarse-to-fine} manner, and enables the Gaussians to traverse further from their initialized locations. After optimization, RAIN-GS applied to random point cloud achieves on-par or even better results compared to 3DGS trained with SfM. By effectively relaxing the requirement of initial point cloud, RAIN-GS opens up new possibilities for applying 3DGS to settings where obtaining accurate initial point cloud is challenging.

\section{Related work}
\label{sec:relwork}
\subsection{Novel view synthesis}
Neural radiance fields (NeRF)~\cite{mildenhall2021nerf} have succeeded in significantly boosting the performance of novel view synthesis by optimizing an MLP that can estimate the density and radiance of any continuous 3D coordinate. With the camera poses of the given images, NeRF learns the MLP by querying dense points along randomly selected rays, that outputs the density and color of each of the queried coordinates. Various follow-ups~\cite{barron2021mip,barron2022mip,du2023learning,hong2023unifying,li2023dynibar,muller2022instant,song2024darf,yang2023freenerf} adopted NeRF as their baseline model and further extend the ability of NeRF to model unbounded or dynamic scenes~\cite{barron2021mip,barron2022mip,li2023dynibar}, lower the required number of images for successful training~\cite{song2024darf,yang2023freenerf}, learn a generalizable prior to resolve the need of per-scene optimization~\cite{du2023learning,hong2023unifying}, or utilize an external hash-grid to fasten the overall optimization process~\cite{muller2022instant}. Although all of these works show compelling results, the volume rendering from dense points along multiple rays makes NeRF hard to apply in real-time settings achieving lower rendering rates of < 1 fps.

Recently, 3D Gaussian splatting (3DGS)~\cite{kerbl20233d} has been recognized as a new solution for the task of novel view synthesis, succeeding in high-quality real-time novel-view synthesis at 1080p resolution with frame rates exceeding  90 fps. 3DGS achieves this by modeling the 3D scene with explicit 3D Gaussians, followed by an efficient and differentiable tile rasterization implemented in CUDA to accelerate the rendering process. Due to the efficient and fast architecture of 3DGS, it gained massive attention from various fields~\cite{luiten2023dynamic, yi2023gaussiandreamer,yang2023deformable3dgs,tang2023dreamgaussian}, extended to model dynamic scenes~\cite{yang2023deformable3dgs,luiten2023dynamic}, or used as an alternative for text-to-3D tasks~\cite{tang2023dreamgaussian,yi2023gaussiandreamer} for fast generation. Nevertheless, the essential requirement of both accurate pose and initial point cloud limits the application of 3DGS to only scenes where SfM can provide accurate point cloud. In this work, we analyze the optimization strategy of 3DGS and present a simple yet effective change in the optimization strategy to expand the versatility of 3DGS to successfully learn the scene without accurate initialization.

\subsection{Structure-from-Motion (SfM)}
SfM techniques~\cite{agarwal2011building,frahm2010building,schonberger2016structure} have been one of the most widely used algorithms to reconstruct a 3D scene. Typical SfM methods output the pose of each input image and a sparse point cloud that includes rough color and position information for each point. These methods employ feature extraction, matching steps, and bundle adjustment. For the task of novel view synthesis, NeRF~\cite{mildenhall2021nerf} utilizes the estimated pose from the SfM and trains an implicit representation to model the scene. The recently proposed 3DGS~\cite{kerbl20233d} utilizes both the accurate pose and initial point cloud as an initialization for the position and color of 3D Gaussians, showing large performance drops when the initial point cloud becomes noisy or not accessible.

Despite the effectiveness of SfM in generating accurately aligned point cloud and precise camera poses, its incremental nature and the computational intensity of the bundle adjustment process significantly increase its time complexity, often to $O(n^4)$ with respect to $n$ cameras involved~\cite{wu2013towards}. Due to the high computational demand, SfM being a hard prerequisite limits the feasibility of NeRF or 3DGS for scenarios that require real-time processing. In addition, SfM often struggles to converge, such as in scenes with symmetry, specular properties, and textureless regions, and when the available views are limited. In these cases, an additional module is required to provide accurate initial point cloud to 3DGS, which can take longer times than SfM~\cite{wang2023dust3r}.
 
\vspace{-5pt}
\section{Preliminary: 3D Gaussian splatting (3DGS)}
\label{sec:preliminary}
Recently, 3DGS~\cite{kerbl20233d} has emerged as a promising alternative for the novel view synthesis task, modeling the scene with multiple 3D Gaussians~\cite{zwicker2002ewa}. 
Each $i$-th Gaussian $G_i$ represents the scene with the following attributes: a position vector $\mu_i\in\mathbb{R}^3$, an anisotropic covariance matrix $\Sigma_i\in\mathbb{R}^{3\times3}$, spherical harmonic (SH) coefficients~\cite{yu2021plenoxels, muller2022instant}, and an opacity logit value $\alpha_i \in [0,1)$. With these attributes, each Gaussian $G_i$ is defined in the world space~\cite{zwicker2002ewa} as follows:
\begin{equation}
    G_i(x) = e^{-\frac{1}{2}(x-\mu_i)^T\Sigma_i^{-1}(x-\mu_i)}.
\end{equation}
To render an image from a pose represented by the viewing transformation $W$, the projected covariance $\Sigma_i'$ is defined as follows:
\begin{equation}
    \Sigma_i' = JW\Sigma_i W^TJ^T,
\end{equation}
where $J$ is the Jacobian of the local affine approximation of the projective transformation. The 2D covariance matrix is simply obtained by skipping the third row and column of $\Sigma_i'$~\cite{zwicker2002ewa}. Finally, to render the color $C(p)$ of the pixel $p$, 3DGS utilizes alpha blending according to the Gaussians depth. For example, when $N$ Gaussians are sorted by depth, the color $C(p)$ is calculated as follows:
\begin{equation}
    C(p) = \sum_{i=1}^{N} c_i \alpha_i G'_i(p)\prod_{j=1}^{i-1} (1 - \alpha_j G'_j(p)),
\end{equation}
where $c_i$ is the view-dependent color value of each Gaussian calculated with the SH coefficients, and $G'_i$ is the 3D Gaussian projected to the 2D screen space.

Since 3DGS starts learning from sparse point cloud, they utilize an adaptive density control to obtain a more accurate and denser representation. By considering if the scene is under-/over-constructed, each of the Gaussians undergoes a clone/split operation. Gaussians that have small covariance are cloned and Gaussians considered to be too large are splitted into two new smaller Gaussians, where the means of new Gaussians are sampled from the probability density function (PDF) of the original Gaussian. When the Gaussian is considered to be too small, the Gaussian is cloned and simply located in the mean of the original Gaussian.
\vspace{-5pt}
\section{Motivation}
\label{sec:motivation}
Although point cloud can be noisy or unavailable in real-world scenarios~\cite{bian2023nope,zhang2022relpose}, 3DGS shows large performance drops depending on the accuracy of the initial point cloud~\cite{kerbl20233d}. To understand the large performance gap of 3DGS, we conduct an in-depth analysis of the original 3DGS optimization scheme and the analysis of SfM initialization in the frequency domain. Our analyses reveal two important characteristics of the optimization scheme of 3DGS: \textbf{1)} the optimization scheme of 3DGS struggles to transport Gaussians from their initialized locations and \textbf{2)} the coarse structure information~(low-frequency components) provided by the accurate initialization enables the adaptive density control method of 3DGS to robustly model the remaining fine details of the scene in a coarse-to-fine manner.

\paragraph{3DGS lacks the ability to transport Gaussians.}
\label{subsec:motivation_3DGS}
To represent and learn the scene with explicit 3D Gaussians, 3DGS first initializes the Gaussians $G_i$ in the world space, whose means $\mu_i$ are defined by the initial point cloud. The point cloud can be either achieved from SfM or initialized randomly.

As mentioned in \cite{charatan23pixelsplat}, the process of fitting a 3DGS model is similar to fitting a Gaussian Mixture Model (GMM), which is well-known for being non-convex and generally solved with the Expectation-Maximization (EM) algorithm \cite{dempster1977maximum}. They further note that, similar to the EM algorithm, training 3DGS from randomly initialized point cloud becomes prone to falling into local minima due to two main reasons. \textbf{1)} The Gaussians can only receive gradients close to their means, mostly from the range not exceeding the distance of a few standard deviations, and \textbf{2)} there is no existing path for the Gaussians that will decrease the loss monotonically. 

Although \cite{charatan23pixelsplat} only analyzes the case of starting from random initialization, we verify that Gaussians can also easily fall into local minima when SfM-initialized point cloud become noisy. As shown in Table~\ref{tab:noise_sfm}, we find that adding a small constant noise or adding a small noise $\epsilon$ sampled from a normal distribution ($\epsilon \sim \mathcal{N}(0,1))$ to the SfM-initialized point cloud, leads to large performance drops. Based on these observations, we hypothesize that the optimization scheme of 3DGS lacks the ability to correct or move the positions of the Gaussians. We empirically verify our hypothesis by calculating the average distance each Gaussian traversed after optimization, as shown in Table~\ref{tab:movement}. It can be seen that the average distance each Gaussian moved is close to zero, indicating that the optimization scheme of 3DGS lacks the ability to move Gaussians, which can lead to the failure of capturing objects located far from the initial positions of the Gaussians. This emphasizes the need for a strategy that can enable the Gaussians to transport further from their initialized locations, in order to successfully train 3DGS from sub-optimal initializations.
\vspace{-5pt}

\begin{table}[t]
\parbox{.42\linewidth}{
  \caption{\textbf{Effect of noise on the initial SfM point cloud.}
  }
  \label{tab:noise_sfm}
  \centering
  \resizebox{1.0\linewidth}{!}{
  \begin{tabular}{l|ccc}
    \toprule
 Init.  & PSNR$\uparrow$ & SSIM$\uparrow$ & LPIPS$\downarrow$  \\
\midrule\midrule
SfM & 27.205 & 0.815 & 0.214  \\
SfM + ${\epsilon}$ & 24.944  & 0.771 & 0.270  \\
SfM + constant& 26.204  & 0.761 & 0.277  \\
\bottomrule
  \end{tabular}
  }
  }
\hfill 
\parbox{.55\linewidth}{
    \caption{\textbf{Analysis of Gaussians movement on Truck scene.}}
    \label{tab:movement}
    \centering
    \resizebox{1.0\linewidth}{!}{
    \begin{tabular}{l|c|c|c}
        \toprule  & 3DGS (SfM) & 3DGS (Random) & Ours (Random) \\
        \midrule \midrule
        Means  & 0.349 &  0.184& 12.100\\
        Stds & 0.726& 0.297& 18.600\\
        Top 1\% & 3.418& 1.524& 82.721\\
        \bottomrule
    \end{tabular}
    }
  }
\end{table}

\paragraph{Accurate initialization guides 3DGS to learn in a coarse-to-fine manner.}
\label{subsec:motivation_SfM}
\begin{figure*}[t]
\begin{center}

\resizebox{1.0\linewidth}{!}
{
    \includegraphics[width=1\linewidth]{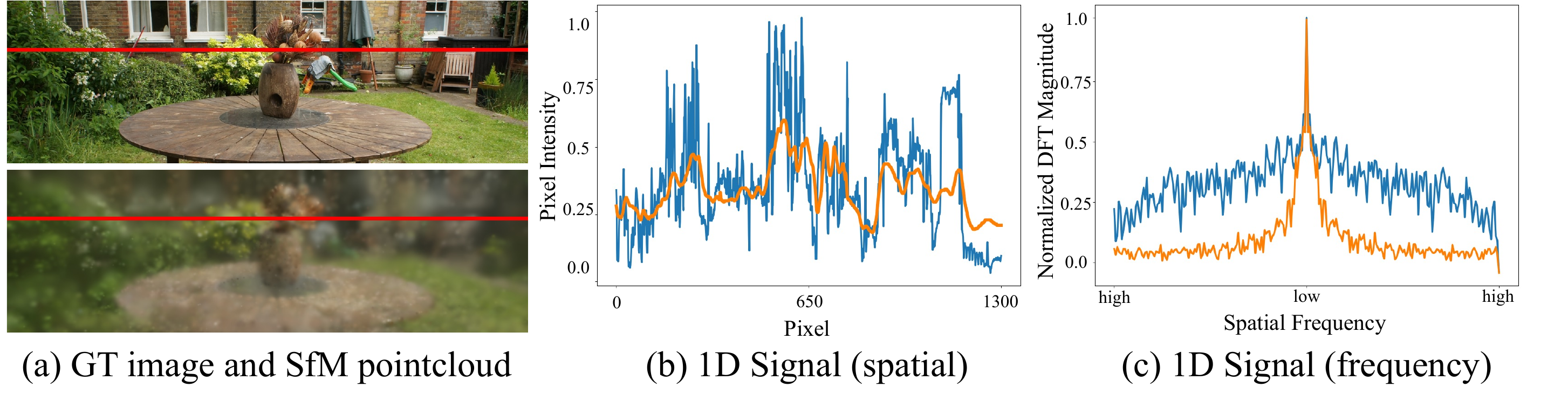}
}
\end{center}
  \vspace{-15pt}
    \caption{\textbf{Analysis of SfM initialization in 3DGS.} (a) The top shows the GT image, and the bottom is the rendered image by 3DGS after only 10 steps with SfM initialization. We can observe that the rendered image is already coarsely-close to GT image. We randomly sample a horizontal line from the image marked in \textcolor{myred}{red}. (b) The pixel intensity along this line are shown, with the GT indicated in \textcolor{myblue}{blue} and the rendered image in \textcolor{myorange2}{orange}. (c) This graph visualizes the magnitude of the frequency components of (b). Since frequencies further from the middle of the x-axis represent high-frequency components, we observe that SfM provides \textit{coarse} approximation of the true distribution.}
    \label{fig:sfm}
\vspace{-15 pt}
\end{figure*}

To further investigate the benefits of SfM initialization, we analyze the rendered images in the frequency domain using Fourier transform~\cite{nussbaumer1982fast}. As shown in Figure~\ref{fig:sfm}, the analysis in the frequency domain demonstrates that SfM initialization provides a coarse approximation of the target distribution. 

As the goal of novel view synthesis is to understand the 3D distribution of the scene, it is necessary to model both low- and high-frequency components of the true distribution. However, prior NeRF frameworks~\cite{lin2021barf, park2021nerfies,yang2023freenerf} argue that NeRF is prone to overfitting and na\"ive optimization leads to over-fast convergence of high-frequency components, expressed with high-frequency artifacts in the rendered image. To circumvent this problem, they adopt a coarse-to-fine learning strategy, which regularizes NeRF to learn the low-frequency components first. Similarly, prior works~\cite{eckart2016accelerated,hertz2020pointgmm} utilizing GMMs for the task of point cloud registration or generation also mention that na\"ive fitting of GMMs can result in converging to local minima. In order to robustly train GMMs, they also adopt a coarse-to-fine strategy, implemented by starting with a small number of Gaussians and recursively increasing the number of total Gaussians. In both NeRFs and GMMs, coarse-to-fine strategy guides the network to learn more robustly, leading to better performance. 

In this perspective, starting the optimization of 3DGS from SfM-initialized point cloud can be understood as benefitting from a similar coarse-to-fine process, where SfM provides the low-frequency components (Figure~\ref{fig:sfm}), and the adaptive density control method of 3DGS adds the Gaussians to learn the remaining high-frequency details. Based on our observations, the success of 3DGS from accurate initialization can be attributed to the low-frequency components guiding the overall training process, preventing the Gaussians from falling into local minima.
This highlights the need for a strategy that can prioritize the learning of the low-frequency components even from sub-optimal initializations, which will then be used to guide the remaining optimization process of 3DGS.
\vspace{-5pt}
\section{Methodology}
\label{sec:method}
\vspace{-5pt}

Following our motivation (Section~\ref{sec:motivation}), we propose a novel optimization strategy, dubbed \textbf{RAIN-GS} (\textbf{R}elaxing \textbf{A}ccurate \textbf{IN}itialization Constraint for 3D \textbf{G}aussian \textbf{S}platting), which enables the Gaussians to prioritize the learning of low-frequency components, and enables the Gaussians to transport further from their initialized locations. Our strategy consists of three key components: \textbf{1)} sparse-large-variance (SLV) initialization (Section~\ref{subsec:sparseinit}), \textbf{2)} progressive Gaussian low-pass filtering (Section~\ref{subsec:lowpass}), and  \textbf{3)} a novel adaptive bound-expanding split (\oursplit) algorithm (Section~\ref{subsec:novel_split}). 

\vspace{-5pt}
\subsection{Sparse-large-variance (SLV) initialization}
\label{subsec:sparseinit}
\vspace{-5pt}

\begin{figure*}[t]
\begin{center}
\resizebox{1.0\linewidth}{!}
{
    \includegraphics[width=1\linewidth]{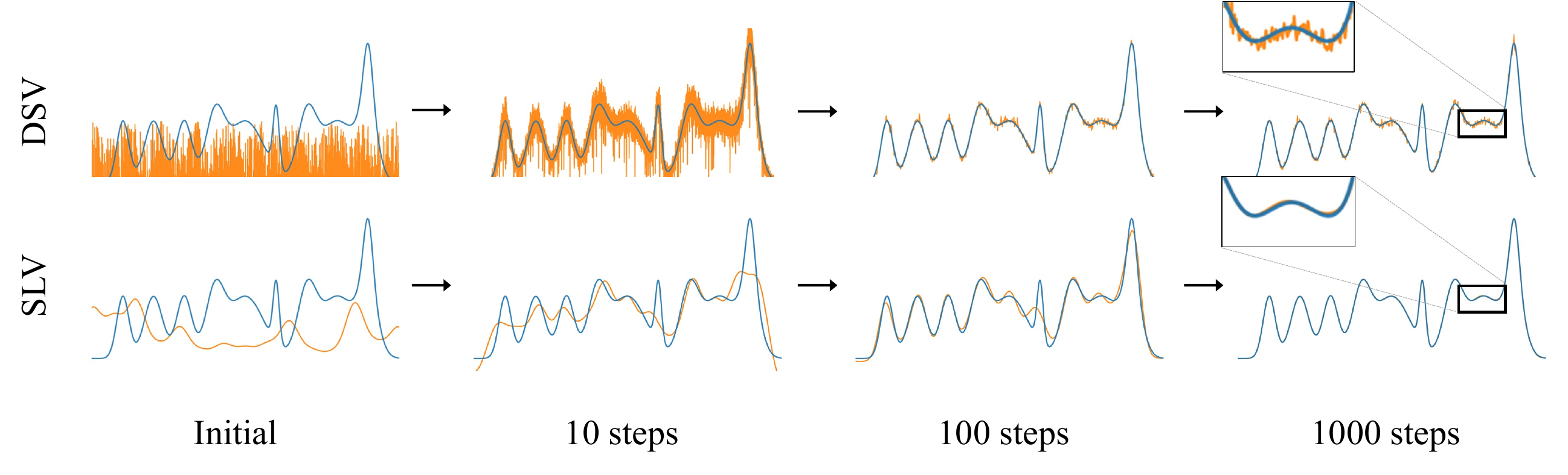}
}
\end{center}
  \vspace{-10pt}
    \caption{\textbf{Toy experiment to analyze different initialization methods.} This figure 
    visualizes the result of our toy experiment predicting the \textcolor{myblue}{target distribution} using a collection of \textcolor{myorange2}{1D Gaussians}, starting from different initialization methods.}
    \label{fig:toy}
\vspace{-10pt}
\end{figure*}

Drawing inspiration from GMMs~\cite{eckart2016accelerated, nichol2022point}, which gradually increase the number of Gaussians to accurately model target point cloud, we observe that the adaptive density control of 3DGS can be viewed as a similar process. Through cloning and splitting operations, 3DGS generally increases the number of Gaussians to find the adequate number of Gaussians required to represent the scene. Based on our findings, we hypothesize that initializing 3DGS with a sparse set of Gaussians will prioritize the learning of low-frequency components, akin to the progressive refinement approach employed by GMMs. This sparse initialization strategy is expected to capture the overall structure of the target point cloud in the early stages of the optimization process, with finer details being added as the number of Gaussians increases.

To verify our hypothesis, we conduct a toy experiment in a simplified 1D regression task. Following the original 3DGS which can be interpreted as the learning process of a 3D target distribution with multiple Gaussians, we use $N$ Gaussians each with learnable means, variances, and weights, which are then blended to model a 1D target signal. Specifically, we follow the initialization methods of 3DGS~\cite{kerbl20233d}, where the means are initialized randomly and the variances are initialized based on the distances of the three nearest neighbors. As a result, sparse initialization of Gaussians leads to a larger initial covariance (SLV) and dense initialization leads to a smaller covariance (DSV). To verify our hypothesis that learning with sparse Gaussians will prioritize the learning of low-frequency components, we conduct our toy experiment using $N=15$ and $N=1000$ for the SLV and DSV initialization respectively. Note that our 1D toy experiment without the adaptive density control method of 3DGS provides a controlled environment isolating the effects of initialization. A detailed explanation of our toy experiment can be found in Section~\ref{supple:C} of the supplementary materials.

As shown in Figure~\ref{fig:toy}, SLV initialization prioritizes the learning of low-frequency components compared to DSV initialization verifying our hypothesis. After 1,000 steps, SLV also shows a better prediction of the target distribution. Similar results can be observed when SLV is applied to 3DGS, as lowering the number of initial Gaussians $N$ in randomly initialized settings significantly improves performance. Following the random initialization method of \cite{kerbl20233d}, which randomly samples point cloud from a scene extent defined as three times the bounding box of the camera poses, SLV prioritizes the learning of low-frequency components, producing fewer high-frequency artifacts. Surprisingly, SLV becomes more effective even until extremely sparse settings~(e.g., as low as $N=10$), verifying the effectiveness of our novel SLV initialization method.

\subsection{Progressive Gaussian low-pass filter control}
\label{subsec:lowpass}
Although our SLV initialization method is effective, we find that after multiple densification steps, the number of 3D Gaussians increases exponentially due to the adaptive density control, which can collapse into similar problems with the DSV initialization. In order to guide the 3D Gaussians to learn in a coarse-to-fine manner and let the 3D Gaussians sufficiently explore the low-frequency components during the early stage of training, we propose a novel progressive control of the Gaussian low-pass filter which is utilized in the rendering stage. 
\vspace{-10pt}
\paragraph{Gaussian low-pass filter for 3DGS.} \label{subsubsec:lowpass} In the rendering stage of 3DGS, the 2D Gaussian $G_{i}'$ projected from a 3D Gaussian $G_i$ is defined as follows:
\begin{equation}
   G_{i}'(x) = e^{-\frac{1}{2}(x-\mu_{i}')^T{\Sigma_{i}'}^{-1}(x-\mu_{i}')}.
\end{equation}
However, directly using projected 2D Gaussians can lead to visual artifacts when they become smaller than the size of a single pixel~\cite{kerbl20233d,yu2023mip}. To ensure coverage of at least one pixel, \cite{kerbl20233d} enlarge the 2D Gaussian's scale by adding a small value to the covariance's diagonal elements as follows:
\begin{equation}
   G_{i}'(x) = e^{-\frac{1}{2}(x-\mu_{i}')^T(\Sigma_{i}'+sI)^{-1}(x-\mu_{i}')},
\end{equation}
where $s$ is a pre-defined value of $s=0.3$ and $I$ is an identity matrix. This process can also be interpreted as the convolution between the projected 2D Gaussian $G_{i}'$ and a Gaussian low-pass filter $h$ (mean $\mu=0$ and variance $\sigma^2 = 0.3$) of $G_{i}' \otimes h$, which is shown to be an essential step to prevent aliasing~\cite{zwicker2002ewa}. After applying convolution with the low-pass filter, the area of the projected Gaussian $G_{i}'$ is approximated by a circle. The radius of this circle is defined by three times the larger eigenvalue from the 2D covariance matrix $(\Sigma_{i}'+sI)$\footnote{We provide a detailed proof in Section B.1 in the supplementary materials.}. 

\paragraph{Progressive low-pass filter control.} 
Instead of using a fixed value of $s$ through the entire optimization process, we notice that this value $s$ can \textbf{ensure} the minimum area each Gaussians have to cover in the screen space. As Gaussians only receive gradients inside the range of a few standard deviations~\cite{charatan23pixelsplat}, learning from wider areas is essential for the Gaussians to learn the coarse structure information of the scene. Therefore, we \textbf{control} $s$ to regularize the Gaussians to cover wider areas during the early stage of training and progressively learn from a more local region. Specifically, as the value $s$ ensures the projected Gaussians area to be larger than $9{\pi}s$\footnote{We provide a detailed proof in Section B.2 in the supplementary materials.}, we define the value $s$ such that $s = HW/9\pi N$, where $N$ indicates the number of Gaussians and $H,W$ indicates the height and width of the image respectively. 

\subsection{Adaptive bound-expanding split (\oursplit) algorithm}
\label{subsec:novel_split}
As discussed in Section~\ref{subsec:motivation_3DGS}, in 3DGS, Gaussians struggle to model scenes far from their initialized locations. Through visualizations of the positions of the Gaussians after training, we empirically find that Gaussians particularly struggle to model scenes \textit{outside} the initial bound defined by the starting point cloud positions (see Section~\ref{subsec:gaussian_distance} of the supplementary materials for visualizations).

This type of behavior is largely attributed to the characteristics of the adaptive density control method of 3DGS~\cite{kerbl20233d}, which applies a clone/split algorithm to the Gaussians. As the new Gaussian is sampled from the PDF of the original Gaussian, it is likely to be located near the original Gaussians, which leads to an increased density of Gaussians near initial locations. Consequently, 3DGS more easily models scenes \textit{inside} the initial bound compared to those \textit{outside} of it, highlighting the need for a strategy that can encourage the Gaussians to move outside the initial bound.

Based on these observations, we propose the adaptive bound-expanding split algorithm, a novel splitting algorithm that can encourage the Gaussians to move outside the initial bound during early steps. Specifically, \oursplit \ algorithm splits each of the Gaussians into three, where two of the Gaussians undergo the same split process as the original split algorithm. However, for the additional third Gaussian, rather than sampling from the PDF of the original Gaussian, we initialize the new location by multiplying a constant value (proportional to the size of the scene extent). Also, we retain the same scale as the original Gaussian without dividing it. This is to encourage the transported Gaussians to receive gradients from larger regions in their new positions. We provide a more detailed explanation of \oursplit \ in Section~\ref{supple:A} in the supplementary materials. 

\vspace{-5pt}
\subsection{Analysis of RAIN-GS}
\vspace{-5pt}
All three key components of our strategy can be interpreted as a component that acts as the ``warm-up'' step for the Gaussians, which guides the Gaussians to first learn the coarse structure of the scene. SLV initialization and progressive Gaussian low-pass filtering facilitate coarse-to-fine optimization whereas \oursplit \ enables the Gaussians to transport to further locations. We demonstrate that our strategy successfully guides 3DGS to learn the coarse structure without any given information (e.g., accurate initial point cloud) which can widen the application of 3DGS to real-world scenarios where the initial point cloud can be noisy or absent. Specifically, as RAIN-GS does not have any restrictions or conditions for the initial point cloud other than being sparse (SLV initialization), we can apply our strategy to both point cloud achieved from SfM~\cite{schonberger2016structure} or random. The effectiveness and versatility of our strategy are further evaluated in Section~\ref{sec:exp}.
\section{Experiments}
\label{sec:exp}
\vspace{-5pt}

\subsection{Experimental Settings}
\vspace{-5pt}
\paragraph{Datasets.} 
To assess the effectiveness of our strategy, we conduct qualitative and quantitative comparisons on the Mip-NeRF360 dataset~\cite{barron2022mip}, Tanks\&Temples dataset~\cite{knapitsch2017tanks}, and Deep Blending dataset~\cite{hedman2018deep} previously utilized in the evaluation of the 3DGS method~\cite{kerbl20233d}. We evaluate the PSNR, LPIPS, and SSIM metrics by constructing a train/test split using every 8th image for testing, as suggested in Mip-NeRF360~\cite{barron2022mip}. We use the same image resolution as 3DGS.

\vspace{-10pt}
\paragraph{Tasks.} As mentioned in Section~\ref{sec:motivation}, SfM can fail in real-world scenarios where the point cloud become noisy or sometimes inaccessible. To assess the robustness of 3DGS~\cite{kerbl20233d} in these scenarios, we first compare our strategy with 3DGS each trained with SfM-initialized point cloud, SfM-initialized point cloud with added noise $\epsilon$ ($\epsilon \sim N(0,1)$), and random point cloud, where our strategy only utilizes random point cloud (Table~\ref{tab:main_mip360}). In addition, to assess the performance of our strategy, we compare our strategy with Plenoxels~\cite{yu2021plenoxels}, InstantNGP-Base, InstantNGP-Big~\cite{muller2022instant}, and 3DGS~\cite{kerbl20233d} (Table~\ref{tab:mip360}). In the latter experiment, we evaluate 3DGS trained with SfM-initialized point cloud, and random point cloud. For our strategy, we train RAIN-GS with SfM-initialized point cloud and random point cloud.

\vspace{-5pt}
\subsection{Implementation details}
\label{subsec:implementation_details}
\vspace{-5pt}
We implement our model based on 3DGS~\cite{kerbl20233d}. We follow the same training process of the existing implementation in all datasets. We apply our strategy to both randomly initialized point cloud and SfM-initialized point cloud. When applying RAIN-GS to randomly initialized point cloud (Table~\ref{tab:main_mip360}, Table~\ref{tab:mip360}), we set the initial number of Gaussians to $N=10$. When applying RAIN-GS to SfM-initialized point cloud (Table~\ref{tab:mip360}), we select a sparse set from the entire point cloud. We either use the top 10\% with the least reprojection error (denoted as <10\%) or use cluster centers (denoted as cluster) obtained through HDBSCAN~\cite{mcinnes2017accelerated}. As RAIN-GS guides the Gaussians to learn the coarse structure of the scene in the early stages of training, we set the first 10,000 steps as the warm-up phase and extend the densification steps by 10,000 steps. During the warm-up phase, the learning rate of the mean $\mu$ is unchanged, and \oursplit\ replaces the original split algorithm. Additional details are included in supplementary materials~\ref{supple:A}.

\begin {table*}[t]
    \caption{\textbf{Quantitative comparison on Mip-NeRF360, Tanks\&Temples and Deep Blending datasets in different initial points conditions}. We compare our method with 3DGS under three different initial point conditions: SfM points, noisy SfM points, and the absence of initial points. 
    }
    \centering
    \resizebox{\textwidth}{!}{
    \begin{tabular}{l|c|ccc|ccc|ccc|ccc|ccc}
        \toprule[0.3ex] \multirow[c]{3}{*}{Methods} &\multirow[c]{3}{*}{ \shortstack{ Init. \\ points }}  & \multicolumn{15}{c}{Mip-NeRF360 Outdoor Scene} \\
        \cline{3-17}
         & & \multicolumn{3}{c|}{bicycle} & \multicolumn{3}{c|}{flowers} & \multicolumn{3}{c|}{garden} & \multicolumn{3}{c|}{stump} & \multicolumn{3}{c}{treehill}\\
         & & PSNR$\uparrow$ & SSIM$\uparrow$ & LPIPS$\downarrow$ & PSNR$\uparrow$ & SSIM$\uparrow$ & LPIPS$\downarrow$ & PSNR$\uparrow$ & SSIM$\uparrow$ & LPIPS$\downarrow$ & PSNR$\uparrow$ & SSIM $\uparrow$ & LPIPS$\downarrow$ &PSNR$\uparrow$ & SSIM $\uparrow$ & LPIPS$\downarrow$  \\
         \midrule \midrule

        \multirow[c]{3}{*}{\shortstack{3DGS~\cite{kerbl20233d} \\ \vspace{-6pt}\tiny }} & {{\shortstack{SfM \\ \tiny }}}
        & \stay{25.246}& \stay{0.771}& \stay{0.205} &  \stay{21.520}&  \stay{0.605}&  \stay{0.336}& \stay{27.410}& \stay{0.868}& \stay{0.103}& \stay{26.550}&  \stay{0.775}& \stay{0.210}&  \stay{22.490}& \stay{0.638}&  \stay{0.317}\\    

         & {{\shortstack{Noisy SfM \\ \vspace{-1.5pt} \tiny }}}
         & \decrease{24.289}{-0.957}& \decrease{0.692}{-0.079}& \decrease{0.307}{+0.102}
         & \decrease{21.028}{-0.492}& \decrease{0.565}{-0.040} & \decrease{0.374}{+0.038}
         & \decrease{26.515}{-0.895}& \decrease{0.846}{-0.022} & \decrease{0.133}{+0.030}
         & \decrease{26.028}{-0.522}& \decrease{0.743}{-0.032} & \decrease{0.252}{+0.042}
         & \decrease{21.728}{-0.762}& \decrease{0.607}{-0.031} & \decrease{0.380}{+0.063}
         \\

         & {{\shortstack{Random \\ \tiny }}}
         &  \decrease{21.034}{-4.212}  & \decrease{0.575}{-25.42} & \decrease{0.378}{+0.173}  
         & \decrease{17.815}{-3.705}  & \decrease{0.469}{-22.48} & \decrease{0.403}{+0.067}  
         & \decrease{23.217}{-4.193}  & \decrease{0.783}{-0.085} & \decrease{0.175}{+0.072} 
         & \decrease{20.745}{-5.805}  & \decrease{0.618}{-0.157} & \decrease{0.345}{+0.135}
         & \decrease{18.986}{-3.504}  & \decrease{0.550}{-0.088} & \decrease{0.413}{+0.096}  
         \\
         
         \midrule
         \hlrow Ours & Random 
         &{25.042}&  {0.747}&  {0.238}&  {21.762}& {0.616}& {0.324}& {26.884}& {0.854}& {0.114}& {26.680}& {0.768}& {0.215}&  {22.528}& {0.621}& {0.342}\\

        \bottomrule[0.3ex]
        \multicolumn{16}{c}{} 
        \\
        \toprule[0.3ex]
        \multirow[c]{3}{*}{Methods} &\multirow[c]{3}{*}{ \shortstack{Init. \\ points }} & \multicolumn{12}{c|}{Mip-NeRF360 Indoor Scene} & \multicolumn{3}{c}{ \multirow{2}{*}{\shortstack{Mip-NeRF360 \\ Average}}} \\
        \cline{3-14}  & & \multicolumn{3}{c|}{room} & \multicolumn{3}{c|}{counter} & \multicolumn{3}{c|}{kitchen} & \multicolumn{3}{c|}{bonsai} &  \\
        
        & & PSNR$\uparrow$ & SSIM$\uparrow$ & LPIPS$\downarrow$ & PSNR$\uparrow$ & SSIM$\uparrow$ & LPIPS$\downarrow$ & PSNR$\uparrow$ & SSIM$\uparrow$ & LPIPS$\downarrow$ & PSNR$\uparrow$ & SSIM $\uparrow$ & LPIPS$\downarrow$ & PSNR$\uparrow$ & SSIM $\uparrow$ & LPIPS$\downarrow$  \\

        \midrule \midrule
          \multirow[c]{3}{*}{\shortstack{3DGS~\cite{kerbl20233d} \\ \vspace{-6pt}\tiny }}  & {{\shortstack{SfM \\ \tiny }}}
          & \stay{30.632}& \stay{0.914}& \stay{0.220}& \stay{28.700}& \stay{0.905}& \stay{0.204}& \stay{30.317}& \stay{0.922}& \stay{0.129}& \stay{31.980}& \stay{0.938}& \stay{0.205}&  \stay{27.205} & \stay{0.815} & \stay{0.214}\\      

          & {{\shortstack{Noisy SfM \\ \vspace{-1.5pt} \tiny }}}
         & \increase{30.740}{+0.108}& \decrease{0.901}{-0.013} & \decrease{0.260}{+0.040}
         & \decrease{23.872}{-4.828}& \decrease{0.838}{-0.067} & \decrease{0.285}{+0.081}
         & \decrease{24.477}{-5.840}& \decrease{0.877}{-0.045} & \decrease{0.175}{+0.046}
         & \decrease{25.823}{-6.157}& \decrease{0.875}{-0.063} & \decrease{0.265}{+0.060}
         & \decrease{24.944}{-2.261} & \decrease{0.771}{-0.044} & \decrease{0.270}{+0.056}
         \\

          & {{\shortstack{Random \\ \tiny }}}
         & \decrease{29.685}{-0.947}  & \decrease{0.894}{-0.020} & \decrease{0.265}{+0.045} 
         & \decrease{23.608}{-5.092}  & \decrease{0.833}{-0.072} & \decrease{0.276}{+0.072}  
         & \decrease{26.078}{-4.239}  & \decrease{0.893}{-0.029} & \decrease{0.161}{+0.032}  
         & \decrease{18.538}{-13.442} & \decrease{0.719}{-0.219} & \decrease{0.401}{+0.196} 
         & \decrease{22.190}{-5.015}  & \decrease{0.704}{-0.111} & \decrease{0.313}{+0.099} \\
         
        \midrule 
        \hlrow Ours & Random &  {30.809}&   {0.906}&  {0.247}&  {28.529}&  {0.895}&  {0.223}&  {31.270}&  {0.920}&  {0.137}&  {31.547}&  {0.934}&  {0.218}&   {27.228}& {0.807}&  {0.229}\\

        \bottomrule[0.3ex]
    \end{tabular}}
    \vspace{3pt}
    \label{tab:main_mip360}
\end{table*}

\begin {table*}[t]
    \vspace{-10pt}
    \centering
    \resizebox{0.9\textwidth}{!}{
    \begin{tabular}{l|c|ccc|ccc|ccc|ccc}
        \toprule[0.3ex]
         \multirow[c]{3}{*}{Methods}&\multirow[c]{3}{*}{ \shortstack{Init. \\ points } }&\multicolumn{6}{c|}{Tanks\&Temples} & \multicolumn{6}{c}{Deep Blending} \\
         \cline{3-14}
        & &\multicolumn{3}{c|}{Truck} & \multicolumn{3}{c|}{Train} & \multicolumn{3}{c|}{DrJohnson} & \multicolumn{3}{c}{Playroom}\\
        & & PSNR$\uparrow$ & SSIM$\uparrow$ & LPIPS$\downarrow$ & PSNR$\uparrow$ & SSIM$\uparrow$ & LPIPS$\downarrow$ & PSNR$\uparrow$ & SSIM$\uparrow$ & LPIPS$\downarrow$ & PSNR$\uparrow$ & SSIM $\uparrow$ & LPIPS$\downarrow$  \\
        \midrule\midrule
        
         \multirow[c]{3}{*}{\shortstack{3DGS~\cite{kerbl20233d} \\ \vspace{-6pt}\tiny }}& {{\shortstack{SfM \\ \tiny }}} & \stay{{25.187}} &  \stay{{0.879}} &\stay{{0.148}}&  \stay{{21.097}} &  \stay{{0.802}}&  \stay{{0.218}}&  \stay{{28.766}}& \stay{{0.899}}&  \stay{{0.244}}&  \stay{30.044}&  \stay{{0.906}}&  \stay{{0.241}} \\
         
           & {{\shortstack{Noisy SfM \\ \vspace{-1.5pt} \tiny }}} & \decrease{23.367}{-1.820}&  \decrease{0.838}{-0.041}& \decrease{0.195}{+0.047}& \decrease{20.212}{-0.885}&  \decrease{0.753}{-0.049}&  \decrease{0.284}{+0.156}& \decrease{27.084}{-1.682}& \decrease{0.882}{-0.017}& \decrease{0.283}{+0.039}& \increase{{30.194}}{+0.150}& \decrease{0.901}{-0.005}& \decrease{0.252}{+0.011}\\
        
          & {{\shortstack{Random \\  \tiny }}} &\decrease{20.149}{-5.038}  & \decrease{0.758}{-0.121} & \decrease{0.248}{+0.100} & \decrease{20.824}{-0.273}  & \decrease{0.772}{-0.030} & \decrease{0.255}{+0.127}  & \decrease{28.668}{-0.098}  & \decrease{0.894}{-0.005} & \decrease{{0.258}}{+0.014} & \decrease{28.358}{-1.686}  & \decrease{0.896}{-0.010} & \decrease{0.258}{+0.017} \\
        \midrule
         \hlrow Ours & Random &{24.816}& {0.865}&  {0.169}&  {21.436}&   {0.786}&  {0.244} &  {28.675}& {0.896}&  0.260&  {30.165}&  {0.903}&  {0.250}\\
        \bottomrule[0.3ex]
    \end{tabular}}
    
    \label{tab:mai_tnt_bl}
\end{table*}

\begin {table*}[t]
    \caption{\textbf{{Quantitative comparison on Mip-NeRF360, Tanks\&Temples and Deep Blending datasets}.} We compare our method with previous approaches, including the random initialization method described in the original 3DGS~\cite{kerbl20233d}. `Cluster' and `<10\%' means our model utilizing clustered SfM points and top 10\% of points with the least reprojection error, respectively.
    }
    \centering
    \resizebox{\textwidth}{!}{
    \begin{tabular}{l|c|ccc|ccc|ccc|ccc|ccc}
        \toprule[0.3ex] \multirow[c]{3}{*}{ Method } & \multirow[c]{3}{*}{ \shortstack{SfM \\ points}} & \multicolumn{15}{c}{Mip-NeRF360 Outdoor Scene} \\
        \cline{3-17}
         & & \multicolumn{3}{c|}{ bicycle } & \multicolumn{3}{c|}{ flowers } & \multicolumn{3}{c|}{ garden } & \multicolumn{3}{c|}{ stump } & \multicolumn{3}{c}{ treehill }\\
        & & PSNR$\uparrow$ & SSIM$\uparrow$ & LPIPS$\downarrow$ & PSNR$\uparrow$ & SSIM$\uparrow$ & LPIPS$\downarrow$ & PSNR$\uparrow$ & SSIM$\uparrow$ & LPIPS$\downarrow$ & PSNR$\uparrow$ & SSIM $\uparrow$ & LPIPS$\downarrow$ &PSNR$\uparrow$ & SSIM $\uparrow$ & LPIPS$\downarrow$  \\

        \midrule \midrule
        
         Plenoxels~\cite{yu2021plenoxels} & \ding{55} & 21.912 & 0.496 & 0.506 & 20.097 & 0.431 & 0.521 & 23.495 & 0.606 & 0.386 & 20.661 & 0.523 & 0.503 & 22.248 & 0.509 & 0.540 \\
        
        INGP-Base~\cite{muller2022instant} & \ding{55}  &{22.193} & 0.491 & 0.487  & {20.348} & 0.450 & 0.481 & {24.599} & 0.649 & 0.312 & {23.626} & 0.574 & 0.450 & {22.364} & 0.518 & 0.489  \\
        
        INGP-Big~\cite{muller2022instant} & \ding{55} &{22.171} & {0.512} & {0.446} & {20.652} & {0.486} & {0.441}  & {25.069} & {0.701} & {0.257} & {23.466} & {0.594} & {0.421} & {22.373}  & {0.542} & {0.450} \\
        
        3DGS~\cite{kerbl20233d}  & \ding{55}  & 21.034  & {0.575} & {0.378}  & 17.815  & {0.469} & {0.403}  & 23.217  & {0.783} & {0.175} & 20.745  & {0.618} & {0.345}  & 18.986 & {0.550} & {0.413}  \\
        3DGS~\cite{kerbl20233d}  &\ding{51}  &  \textbf {25.246}&  \textbf {0.771}&  \textbf {0.205}&  21.520&  0.605&  0.336&  \textbf {27.410}&  \textbf {0.868}&  \textbf {0.103}&  26.550& \underline {0.775}&  0.210&  22.490&  \textbf {0.638}&  \textbf {0.317}\\

        \midrule
        \hlrow Ours &\ding{55}  & 25.042&  0.747& 0.238&  \underline {21.762}&  0.616&  0.324&   26.884&  0.854&  0.114&  26.680&  0.768&  0.215&   22.528&  0.621&  0.342\\
        \hlrow Ours & cluster &  25.211&    0.767&   0.211&   21.704&  \underline {0.617}&  \underline {0.321}&   \underline {27.166}&  \underline {0.862}&  \textbf {0.103}&  \underline {26.816}&  \underline {0.775}&  \underline {0.204}&  \underline {22.589}&   0.630&   0.328\\
        \hlrow Ours &<10\%  & \underline {25.214}&  \underline {0.769}& \underline {0.208}&  \textbf {21.858}&  \textbf {0.619}&  \textbf {0.320}&   27.130&  0.861& \underline {0.107}&  \textbf {26.899}&  \textbf {0.777}&  \textbf {0.202}&  \textbf {22.618}& \underline {0.632}& \underline {0.327}\\

        \bottomrule[0.3ex]
        \multicolumn{17}{c}{}
        \\
        \toprule[0.3ex]
        \multirow[c]{3}{*}{ Method } & \multirow[c]{3}{*}{ \shortstack{SfM \\ points}} & \multicolumn{12}{c|}{Mip-NeRF360 Indoor Scene} & \multicolumn{3}{c}{ \multirow{2}{*}{\shortstack{Mip-NeRF360 \\ Average}}} \\
        \cline{3-14} & & \multicolumn{3}{c|}{ room } & \multicolumn{3}{c|}{ counter } & \multicolumn{3}{c|}{ kitchen} & \multicolumn{3}{c|}{ bonsai } & \\
        
        & & PSNR$\uparrow$ & SSIM$\uparrow$ & LPIPS$\downarrow$ & PSNR$\uparrow$ & SSIM$\uparrow$ & LPIPS$\downarrow$ & PSNR$\uparrow$ & SSIM$\uparrow$ & LPIPS$\downarrow$ & PSNR$\uparrow$ & SSIM $\uparrow$ & LPIPS$\downarrow$ & PSNR$\uparrow$ & SSIM $\uparrow$ & LPIPS$\downarrow$  \\

        \midrule \midrule 
        Plenoxels~\cite{yu2021plenoxels} & \ding{55}  &27.594 & 0.842 & 0.419 & 23.624 & 0.759 & 0.441 & 23.420 & 0.648 & 0.447 & 24.669 & 0.814 & 0.398 & 23.080 & 0.625 & 0.462\\
        
        INGP-Base~\cite{muller2022instant} & \ding{55}  &29.269 & 0.855 & 0.301 & {26.439} & 0.798 & 0.342 & {28.548} & 0.818 & 0.254  & {30.337} & {0.890} & {0.227} & {25.302} & 0.671 & 0.371  \\
        
        INGP-Big\cite{muller2022instant} &\ding{55}  & {29.690} & {0.871} & {0.261} & {26.691} & {0.817} & {0.306} & {29.479} & {0.858} & {0.195} & {30.685} & {0.906} & {0.205} & {25.586} & {0.699} & {0.331} \\
        

        3DGS~\cite{kerbl20233d}  & \ding{55}  & {29.685}& {0.894} & {0.265} & 23.608 &  {0.833} &  {0.276}  & 26.078 & {0.893} & {0.161}  & 18.538 & 0.719 & 0.401 & 22.190 & {0.704} & {0.313} \\
        3DGS~\cite{kerbl20233d} & \ding{51}  &  30.632& \textbf {0.914}&  \textbf {0.220}& \underline {28.700}&  \textbf {0.905}&  \textbf {0.204}&  30.317&  \underline {0.922}&  \textbf {0.129}&  \textbf {31.980}&  \textbf {0.938}&  \textbf {0.205}&  27.205 &  \textbf {0.815}&  \textbf {0.214}\\

        \midrule
        
        \hlrow Ours & \ding{55}  & \underline {30.809}&   0.906&  0.247&  28.529&  0.895&  0.223& \underline {31.270}&  0.920&  0.137&   31.547&  0.934&  0.218&   27.228&  0.807&  0.229\\
        \hlrow Ours &cluster&30.363&   0.909&   0.233&  28.594&  0.901&   0.211&  31.136& \textbf {0.923}&  \underline {0.133}&   31.695&   \underline{0.935}&   0.217&  \underline {27.253}&  \underline {0.813}&   0.218\\
        \hlrow Ours & <10\%  & \textbf {30.843}& \underline {0.912}& \underline {0.230}&   \textbf {28.703}& \underline {0.902}& \underline {0.209}&  \textbf {31.364}&  \underline {0.922}& \underline {0.133}&  \underline {31.829}&  \textbf {0.938}&  \underline {0.214}&   \textbf {27.384}&  \textbf {0.815}& \underline {0.217}\\

        \bottomrule[0.3ex]
    \end{tabular}}
    \vspace{2pt}

    \label{tab:mip360}
\end{table*}

\begin {table*}[t]
    \vspace{-10pt}
    \centering
    \resizebox{0.9\textwidth}{!}{
    \begin{tabular}{l|c|ccc|ccc|ccc|ccc}
        \toprule[0.3ex]
         \multirow[c]{3}{*}{ Methods }&\multirow[c]{3}{*}{ \shortstack{SfM \\ points} }&\multicolumn{6}{c|}{Tanks\&Temples} & \multicolumn{6}{c}{Deep Blending} \\
         \cline{3-14}
         &&\multicolumn{3}{c|}{ Truck } & \multicolumn{3}{c|}{ Train } & \multicolumn{3}{c|}{ DrJohnson } & \multicolumn{3}{c}{ Playroom }\\
        && PSNR$\uparrow$ & SSIM$\uparrow$ & LPIPS$\downarrow$ & PSNR$\uparrow$ & SSIM$\uparrow$ & LPIPS$\downarrow$ & PSNR$\uparrow$ & SSIM$\uparrow$ & LPIPS$\downarrow$ & PSNR$\uparrow$ & SSIM $\uparrow$ & LPIPS$\downarrow$  \\
        \midrule \midrule
        
        
        Plenoxels~\cite{yu2021plenoxels} & \ding{55} &23.221 & 0.774 & 0.335 & 18.927 & 0.663 & 0.422 & 23.142 & 0.787 & 0.521 & {22.980} & {0.802} & 0.465 \\
        
        INGP-Base~\cite{muller2022instant} & \ding{55} & {23.260} & {0.779} & 0.274  & 20.170 & 0.666 & 0.386 & 27.750 & 0.839 & 0.381 & 19.483 & 0.754 & 0.465   \\
        
        INGP-Big~\cite{muller2022instant} &\ding{55} & {23.383} & {0.800} & {0.249} & {20.456}& {0.689} & {0.360}  & {28.257} & {0.854} & {0.352} & 21.665 & 0.779 & {0.428} \\
        
        3DGS~\cite{kerbl20233d} &\ding{55} & 21,149  & 0.758 & {0.248} & {20.824}  & {0.772} & {0.255}  & {28.668}  & {0.894} & \underline{0.258} & {28.358}  & {0.896} & {0.258}  \\

        3DGS~\cite{kerbl20233d} &\ding{51} & \underline{25.187}& \textbf{0.879}& \textbf{0.148}&  21.097& \textbf{0.802}& \textbf{0.218}& \textbf{28.766}& \textbf{0.899}& \textbf{0.244}&  30.044& \textbf{0.906}& \textbf{0.241}\\
        \midrule
        
        \hlrow Ours &\ding{55} &  24.816& 0.865 & 0.169& \underline{21.436}&   0.786&  0.244& \underline{28.675}& \underline{0.896}&  0.260& \textbf{30.165}& \underline{0.903}&  0.250\\
        \hlrow Ours &cluster&  24.893&  0.869&  0.160&  21.180&   0.786&  0.243& 28.529& 0.895&  0.260& 29.994&  0.902&  0.250\\
        \hlrow Ours & <10\% & \textbf{25.190}& \underline{0.873}& \underline{0.159}& \textbf{21.741}& \underline{0.796}& \underline{0.235}& 28.567& \underline{0.896}& 0.261& \underline{30.065}&  0.902& \underline{0.249}\\

        \bottomrule[0.3ex]
    \end{tabular}}
    \vspace{-10pt}
    \label{tab:tnt_bl}
\end{table*}

\begin{figure}[!t]
\begin{center}
{
    \includegraphics[width=1\linewidth]{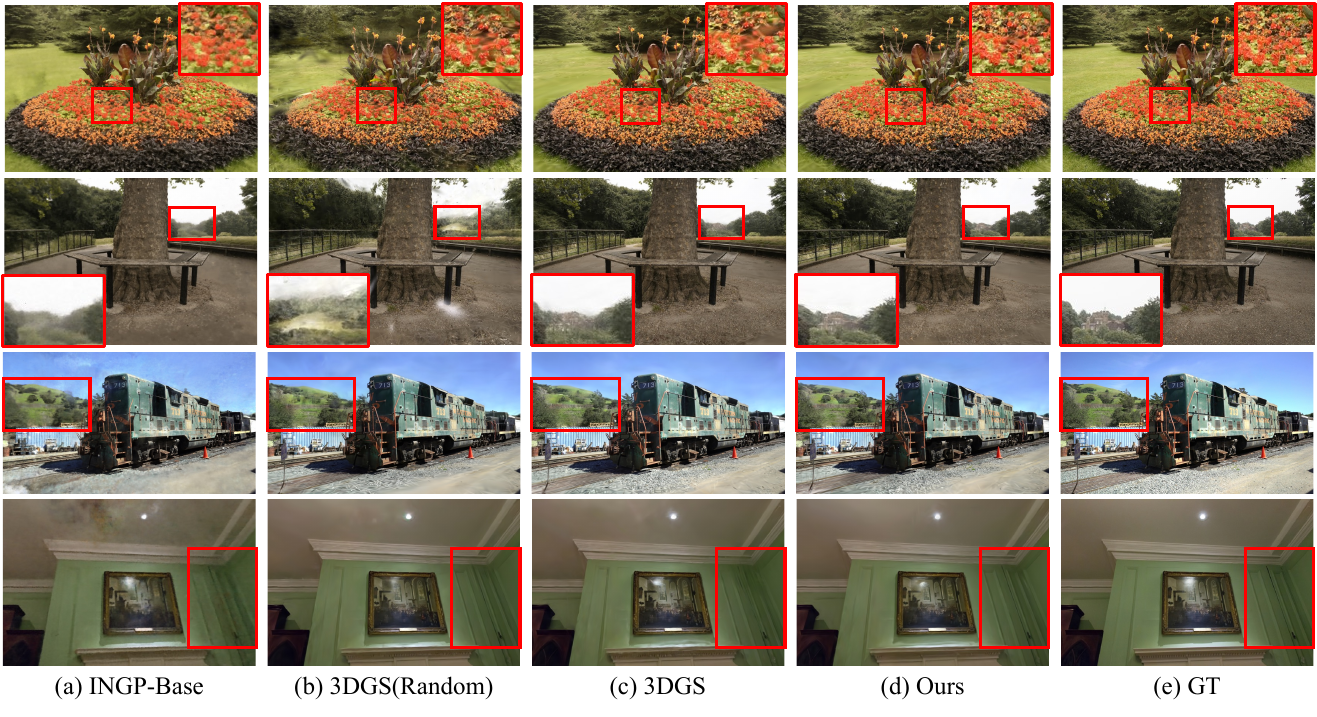}
}
\end{center}
  \vspace{-15pt}
    \caption{\textbf{Qualitative results on Mip-NeRF360, Tanks\&Temples, and Deep Blending dataset.}}
    \label{fig:main_qual}

\vspace{-10pt}
\end{figure}

\subsection{Quantitative comparison with 3DGS in real-word scenarios}
\vspace{-5pt}
As mentioned above, SfM might fail or produce noisy point cloud in real-world scenarios. To assess the robustness of 3DGS~\cite{kerbl20233d} in these cases, we evaluate 3DGS trained with SfM points, noisy SfM points, and random points. The results are compared with ours trained with random points.

As shown in Table~\ref{tab:main_mip360}, 3DGS is highly dependent on the accuracy of initial point cloud, showing large performance drops when trained with noisy SfM point cloud (denoted as Noisy SfM) and random point cloud (denoted as Random). In contrast, our strategy trained from random point cloud shows on-par or even better results compared to 3DGS trained with accurate SfM point cloud.

\vspace{-5pt}
\subsection{Qualitative and quantitative comparison}
\vspace{-5pt}
In Table~\ref{tab:mip360}, we compare our model with Plenoxels~\cite{yu2021plenoxels}, InstantNGP-Base, InstantNGP-Big~\cite{muller2022instant}, and 3DGS~\cite{kerbl20233d} on the Mip-NeRF360 dataset~\cite{barron2022mip}, Tanks\&Temples dataset~\cite{knapitsch2017tanks}, and Deep Blending dataset~\cite{hedman2018deep}. All NeRF methods~\cite{yu2021plenoxels,muller2022instant,barron2021mip} are trained without SfM points. We evaluate 3DGS~\cite{kerbl20233d} and ours trained with SfM-initialized and randomly initialized point cloud. For ours trained with SfM-initialized point cloud, we evaluate utilizing different clustering methods of selecting the top 10\% with the least reprojection error (denoted as <10\%) or using cluster centers (denoted as cluster) obtained through HDBSCAN~\cite{mcinnes2017accelerated}.

In Figure~\ref{fig:main_qual}, we provide qualitative results on novel test views. We compare with InstantNGP-Base ((a)), 3DGS trained with random point cloud ((b)) and SfM initialized point cloud ((c)), and ours trained with random point cloud ((d)). By comparing (b) and (d), we demonstrate the effectiveness of our strategy, which removes any unwanted high-frequency artifacts or floaters and also successfully captures scene objects located far away from the viewpoint. By comparing (c) and (d), ours trained with random point cloud show better results than 3DGS trained with SfM initialized point cloud, capturing fine details more effectively and rendering the background more accurately. More qualitative images are included in Section~\ref{supple:E} in the supplementary materials.

\subsection{Ablation studies}
\label{subsec:ablation_details}

\begin{wraptable}{r}{0.5\linewidth}
    \centering
    \vspace{-10pt}
    \caption{\textbf{Ablation  on core components}}
    \label{tab:c2f_ablation}
    \vspace{-5pt}
    \resizebox{\linewidth}{!}{
    \begin{tabular}{c|c|c|ccc}
    \toprule
    Low-pass filter& Init. & \oursplit &PSNR$\uparrow$ & SSIM$\uparrow$ & LPIPS$\downarrow$\\
    \midrule\midrule
    Constant &  DSV &  \ding{55} & 22.190	& 0.704 & 0.313 \\
    Constant &  SLV &  \ding{55} & 25.675 & 0.749 & 0.288 \\
    Ours & SLV &\ding{55}  & 26.180 & 0.761 & 0.280\\
    \hlrow \textbf{Ours} & \textbf{SLV}&\ding{51}&  \textbf{27.228} & \textbf{0.807} & \textbf{0.229}\\
    \bottomrule

    \end{tabular}}
    \vspace{-10pt}
\end{wraptable}

In Table~\ref{tab:c2f_ablation}, we validate the effectiveness of each component in our method trained in the Mip-NeRF360 dataset~\cite{barron2022mip} with randomly initialized point cloud. The comparison of Sparse-large-variance (SLV) and dense-small-variance (DSV) initialization is done by using $N=10$ and $N=1,000,000$ respectively. We compare our progressive Gaussian low-pass filter control method with using a constant low-pass filter value ($s=0.3$) as done in the original 3DGS~\cite{kerbl20233d}. We also compare utilizing our novel \oursplit \ method and the original split algorithm utilized in 3DGS. All three ablations show that using our components achieves significantly better results and using all three key components achieves the best results. More detailed and various ablation studies can be found in Section~\ref{supple:ablation} in the supplementary materials.

\section{Conclusion}
\label{sec:conclusion}
In this work, we introduce \textbf{RAIN-GS} (\textbf{R}elaxing \textbf{A}ccurate \textbf{IN}itialization Constraint for 3D \textbf{G}aussian \textbf{S}platting), a novel optimization strategy that effectively removes the constraint for accurate initialization. By combining sparse-large-variance (SLV) initialization, progressive Gaussian low-pass filter control, and the adaptive bound-expanding split (\oursplit) algorithm, our strategy successfully guides 3D Gaussians to prioritize the learning of low-frequency components and encourages the Gaussians to transport to further locations from the initialized locations. RAIN-GS effectively removes the strict reliance on accurate point cloud achieved from Structure-from-Motion (SfM), opening up new possibilities for 3DGS in scenarios where achieving accurate point cloud is challenging.


\clearpage
\appendix

\section*{Supplementary materials}

\definecolor{mygray}{gray}{0.9}

In the supplementary materials, we provide a more detailed analysis of our experiments and implementation details, together with additional results of rendered images using our strategy. Specifically, in Section~\ref{supple:A}, we provide the implementation details of how our \textbf{RAIN-GS} (\textbf{R}elaxing \textbf{A}ccurate \textbf{IN}itialization Constraint for 3D \textbf{G}aussian \textbf{S}platting) can be applied to original 3D Gaussian splatting (3DGS)~\cite{kerbl20233d}. In Section~\ref{supple:B}, we give a more detailed explanation of the convolution process with the Gaussian low-pass filter used in the rendering stage and how we choose the low-pass filter value $s$ (diagonal values of the covariance matrix $sI$ of the Gaussian low-pass filter) for our progressive Gaussian low-pass filter control. In Section~\ref{supple:C}, we explain the details of our toy experiment shown in Section~\ref{subsec:sparseinit} of our main paper: learning to regress 1D signal with multiple 1D Gaussians. In Section~\ref{supple:D}, we provide additional analysis including comparisons of computational resources with different methods and limitations of our strategy. In Section~\ref{supple:E}, we provide additional qualitative results. In Section~\ref{supple:F}, we show how our strategy can be applied to training 3DGS with a limited number of images.

\section{Implementation Details}
\label{supple:A}
\subsection{RAIN-GS}
We implement our strategy based on the official code of 3DGS~\cite{kerbl20233d}. As mentioned in Section ~\ref{subsec:implementation_details} of our main paper, except for the hyperparameters for increasing spherical harmonic (SH) degrees and 3D Gaussian scale division ratio, we follow the original implementation details and hyperparameters of 3DGS~\cite{kerbl20233d}. We start increasing SH degree every 1,000 steps after 5,000 steps and we lower 3D Gaussian scale division ratio 1.6 to 1.4.
Our novel strategy consists of three components: \textbf{1)} sparse-large-variance (SLV) random initialization, \textbf{2)} progressive Gaussian low-pass filter control and \textbf{3)} \oursplit. 
While 3DGS begins with the SFM point cloud which provides a coarse approximation of the scene, RAIN-GS starts with significantly fewer Gaussians and without any coarse information. Consequently to achieve an initial coarse approximation, it requires more time and steps to densify the Gaussians. Thus we add warm-up phase in training and our three main components are only applied during the warm-up phase. Specifically, the traditional 3DGS training can be divided into two main phases: (1) densification and (2) optimization. The densification phase typically involves about 15,000 steps, followed by 15,000 steps of optimization. The learning rate of the mean $\mu$ start with $1.6e^{-4}$ and exponentially decay to $1.6e^{-6}$ during densification phase. For RAIN-GS, however, the optimization phase is reduced to 5,000 steps, and 10,000 steps of warm-up phase is introduced prior to densification phase. While warm-up phase, the learning rate of the mean $\mu$ is kept $1.6e^{-4}$ and our componets are applied to help the Gaussians learn a coarse approximation. Then, the remaining process follows the same learning process as the original 3DGS. 

\subsubsection{Sparse-large-variance (SLV) random initialization} 
\begin{figure*}[ht]
\begin{center}
\resizebox{1.0\linewidth}{!}
{
    \includegraphics[width=1\linewidth]{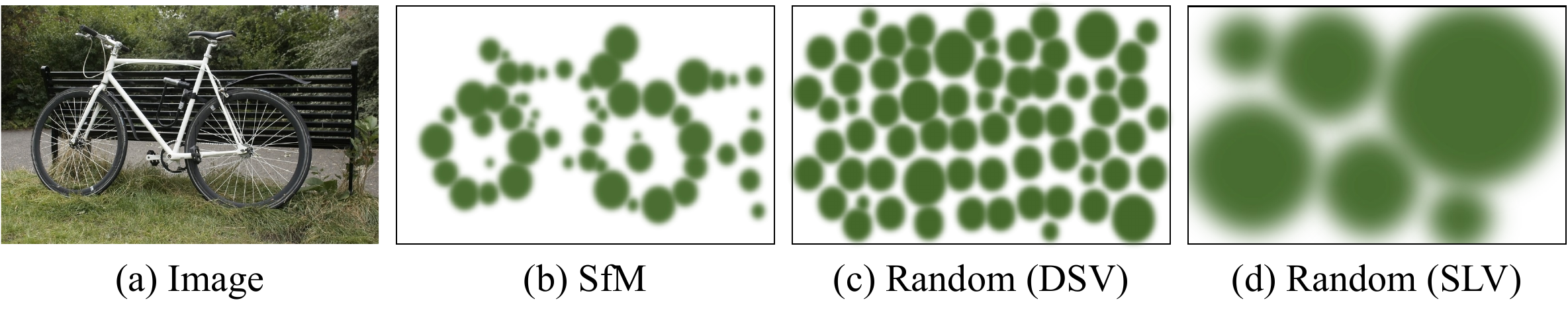}
}
\end{center}
    \caption{\textbf{Visualization of different initialization methods.} This figure illustrates the effect of different initialization methods. \textbf{(a)} Ground truth image. \textbf{(b)} Initialized point cloud from Structure-from-Motion (SfM). \textbf{(c)} Dense-small-variance (DSV) random initialization with small initial covariances due to the short distance between Gaussians. \textbf{(d)} Sparse-large-variance (SLV) random initialization with large initial covariances due to wider distance between Gaussians.}
    \label{fig:init}
\end{figure*}

Following the original implementation of 3DGS~\cite{kerbl20233d}, we initialize random point cloud within the boundary defined as three times the size of the camera's bounding box. Instead of the original dense-small-variance random initialization where the initial number of Gaussians $N$ is set as $N > 100K$, we only initialize 10 Gaussians as $N = 10$. As the initial covariance of 3D Gaussian is defined based on the distances of the three nearest neighbors, sparse initialization leads to a larger initial covariance, resolving the requirement for additional modification. This simple change of code (expressed with a \colorbox{mygray}{gray box} in Algorithm~\ref{alg:init}) largely improves the performance of 3DGS based on our analysis of sparse-large-variance initialization. An illustrations of SfM initialization, dense-small-variance (DSV) initialization and sparse-large-variance (SLV) initialization are shown in Figure~\ref{fig:init}.

\algnewcommand\algorithmicforeach{\textbf{for each}}
\algdef{S}[FOR]{ForEach}[1]{\algorithmicforeach\ #1\ \algorithmicdo}

\begin{algorithm}
\caption{Sparse-large-variance (SLV) random initialization}
\label{alg:init}
\begin{algorithmic}[1]
    \State \colorbox{mygray}{$N \gets$ Number of initial Gaussians (e.g., $N=10$)}
    \State $P \gets$ Cameras
    \State $D \gets$ MaxCameraDistance($P$)
    \State $i \gets 0$	\Comment{Iteration Count}
    \While{$i < N$}
    \State $\mu \gets$ RandomCubicSample(3 $\times$ D)
    \State $\Sigma \gets$ AvgNeighborDistance()
    \State $c \gets$ RandomInit()
    \State $\alpha \gets$ InverseSigmoid(1)
    \State $(M, S, C, A) \gets$ ($\mu, \Sigma, c, \alpha$) \Comment{Appends}
    \State $i \gets i + 1$
    \EndWhile
    \State return $(M, S, C, A)$
\end{algorithmic}
\end{algorithm}

\subsubsection{Progressive Gaussian low-pass filter control} In the original implementation of 3DGS, the Gaussian low-pass filter is used in the rendering stage to ensure the projected 2D Gaussians to cover at least one pixel in the screen space. To enlarge the 2D Gaussians, 3DGS uses a fixed Gaussian low-pass filter (mean $\mu=0$ and variance $\sigma^2 = s = 0.3$). Instead of using a fixed sigma value $\sigma$ for this low-pass filter, we propose a progressive Gaussian low-pass filter control, where the sigma value starts with a large value and progressively reduces to $\sigma^2 = 0.3$. This can be efficiently implemented. Instead of passing a fixed value of $s=0.3$ as the diagonal values of the covariance matrix $sI$ of the Gaussian low-pass filter, we pass a progressive value adaptively defined with the image height, width, and number of Gaussians taken into account as $\text{min(}\text{max(}HW/9\pi N, 0.3\text{), 300.0)}$ every 1,000 steps. Our implementation is expressed in line 8 of Algorithm~\ref{alg:low_pass}.

\begin{algorithm}
    \caption{Progressive Gaussian low-pass filter control \& \oursplit}
    \label{alg:low_pass}
    \begin{algorithmic}[1]
			\State $M \gets$ SfM Points	\Comment{Positions}
			\State $S, C, A \gets$ InitAttributes() \Comment{Covariances, Colors, Opacities}
                \State $s \gets$  Scene Extent
                \State $B_0 \gets$ Center Point of Bounding Box
			\State $i \gets 0$	\Comment{Iteration Count}

			\While{not converged}

			\State $V, \hat{I}, H, W \gets$ SampleTrainingView()	\Comment{Camera $V$, Image, Height and Width}
			\If{LowPassRefinementIteration(i)}
                \State 
                \colorbox{mygray}{$h \gets \text{Min(}\text{Max(}HW/9\pi N, 0.3\text{), 300)}$}  \Comment{Progressive low-pass filter control}
                \EndIf
			\State $I \gets$ Rasterize($M$, $S$, $C$, $A$, $V$, $h$)	\Comment{Rasterization with low-pass filter}
			
			\State $L \gets Loss(I, \hat{I}) $ \Comment{Loss}
			
			\State $M$, $S$, $C$, $A$ $\gets$ Adam($\nabla L$) \Comment{Backprop \& Step}

			\If{IsRefinementIteration($i$)}
			\ForAll{Gaussians $(\mu, \Sigma, c, \alpha)$ $\textbf{in}$ $(M, S, C, A)$}
			\If{$\alpha < \epsilon$ or IsTooLarge($\mu, \Sigma)$}	\Comment{Pruning}
			\State RemoveGaussian()	
			\EndIf
			\If{$\nabla_p L > \tau_p$} \Comment{Densification}
			\If{$\|S\| > \tau_S$}	\Comment{Over-reconstruction}
			\State SplitGaussian($\mu, \Sigma, c, \alpha$)
            \State  \colorbox{mygray}{$\mu' \gets \lambda_s s \cdot (\mu - B_0) + B_0 $}
			\State \colorbox{mygray}{CloneGaussian($\mu', \Sigma, c, \alpha$) }\Comment{\oursplit}
			\Else								\Comment{Under-reconstruction}
			\State CloneGaussian($\mu, \Sigma, c, \alpha$)
			\EndIf	
			\EndIf
			\EndFor		
			\EndIf
			\State $i \gets i+1$
			\EndWhile
    \end{algorithmic}
\end{algorithm}

\subsubsection{\oursplit}
In addition to the split of 3DGS, the \oursplit \ generates an additional Gaussian and transport it to a distant location. Specifically, new position is determined by calculating the vector between the center of bounding box and the original Gaussian position, then multiplying it by 0.3 times the scene extent. The pseudo code for this process is provided in line 21 and 22 of Algorithm~\ref{alg:low_pass}.
Due to our SLV initialization and progressive Gaussian low-pass filtering, the Gaussians are sparse and have large covariance during the early steps of training, resulting in a majority of the Gaussians only undergoing the split process. Thus, the \oursplit \ can transport a sufficient number of Gaussians in early stage. 

\clearpage
\section{{Proof}}
\label{supple:B}

\begin{figure*}[!t]
\begin{center}
\resizebox{1.0\linewidth}{!}
{
    \includegraphics[width=1\linewidth]{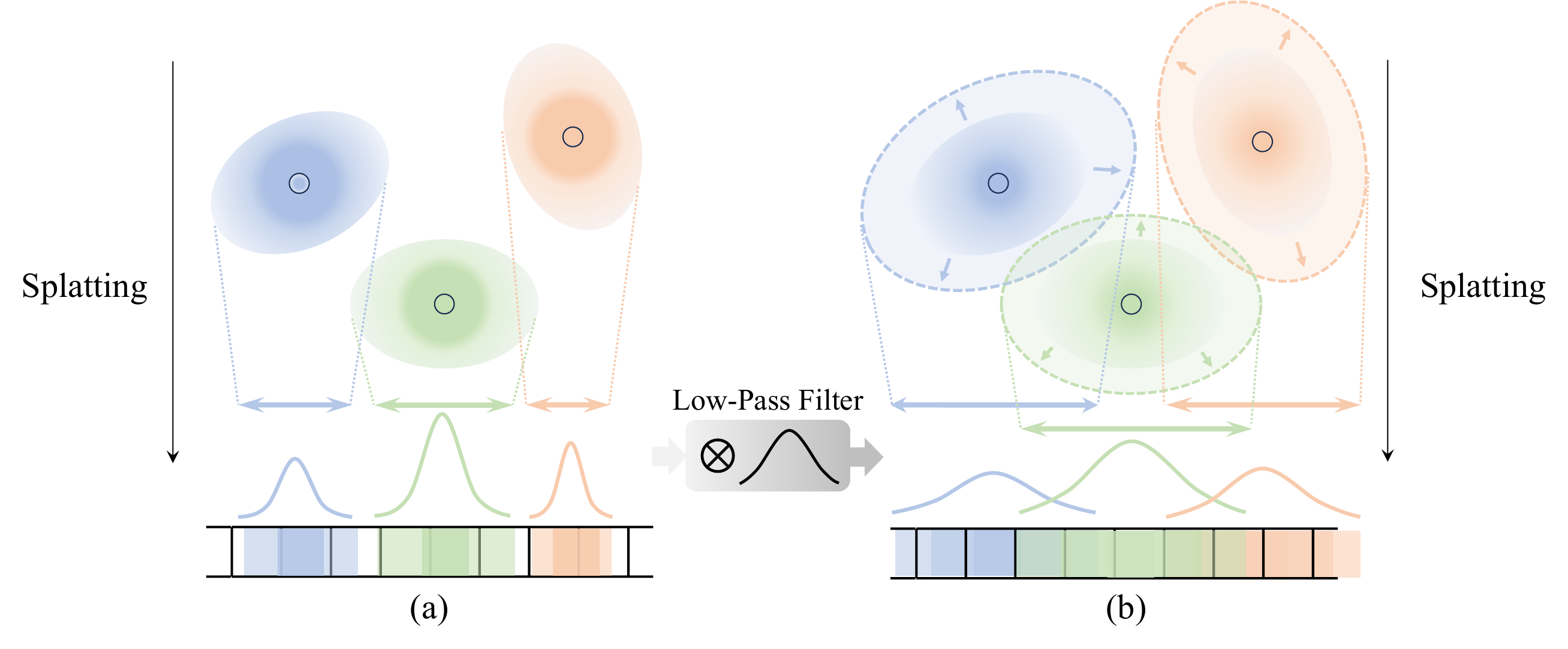}
}
\end{center}
    \caption{\textbf{Visualization of low-pass filter.} This figure shows the visualization of the effect of the low-pass filter. As shown in (b), the convolution of the splatted 2D Gaussian with the low-pass filter expands the area the Gaussian is splatted onto, resulting in the Gaussians affecting larger areas than na\"ive splatting as shown in (a).}
    \label{fig:lowpass}
\end{figure*}

\subsection{Proof on radius of a Gaussian convolved with a low-pass filter}
As mentioned in Section~\ref{subsubsec:lowpass} of our main paper, the 3D Gaussians $G_i$ is projected to 2D Gaussians $G_{i}'$ in the screen space as follows:
\begin{equation}
   G_{i}'(x) = e^{-\frac{1}{2}(x-\mu_{i}')^T{\Sigma_{i}'}^{-1}(x-\mu_{i}')}.
\end{equation}
To ensure the 2D Gaussian $G_{i}'$ to cover at least one pixel, 3DGS adds a small value $s$ to the diagonal elements of the 2D covariance $\Sigma_{i}'$ as follows:
\begin{equation}
   G_{i}'(x) = e^{-\frac{1}{2}(x-\mu_{i}')^T(\Sigma_{i}'+sI)^{-1}(x-\mu_{i}')},
\end{equation}
where $I$ is the $2\times2$ identity matrix. This process can be understood as the convolution between the 2D Gaussian $G_{i}'$ and the Gaussian low-pass filter $h$ (mean $\mu=0$ and variance $\sigma^2 = s = 0.3$) of $G_{i}' \otimes h$. This is due to the nature of Gaussians where the convolution of Gaussians with the variance matrices $V$ and $Z$ results in a Gaussian with the variance matrix $V+Z$ as follows:
\begin{equation}
    G_{1}(x) = e^{-\frac{1}{2}(x-\mu_{i})^T{V}^{-1}(x-\mu_{i})} \quad G_{2}(x) = e^{-\frac{1}{2}(x-\mu_{i})^T{Z}^{-1}(x-\mu_{i})},
\end{equation}
\begin{equation}
    (G_{1} \otimes G_{2})(x) = e^{-\frac{1}{2}(x-\mu_{i})^T({V+Z})^{-1}(x-\mu_{i})}.
\end{equation}
Following the convolution process, 3DGS estimates the projected 2D Gaussian's area to identify its corresponding screen tiles. This is done by calculating $k$ times the square root of the larger eigenvalue of $(\Sigma_{i}'+sI)$, which represents the radius of the approximated circle, and $k$ is the hyperparameter that determines the confidence interval of the 2D Gaussian. Figure~\ref{fig:lowpass} illustrates the low-pass filter’s effect, where the projected Gaussian is splatted to wider areas in (b) compared to (a).

\subsection{Proof on progressive low-pass filter size}
In Section~\ref{subsubsec:lowpass} of our main paper, we define the value $s$ for our progressive Gaussian low-pass filter control based on the fact that the area of the projected 2D Gaussians is at least $9{\pi}s$. As the area of the projected 2D Gaussian is defined as the circle whose radius is $k$ times the square root of the larger eigenvalue of $(\Sigma_{i}'+sI)$, we have to first calculate the eigenvalues of $(\Sigma_{i}'+sI)$. If we define the eigenvalues of $\Sigma_{i}$ as $\lambda_{i1},\lambda_{i2}$, since the eigenvalue of $sI$ is $s$, the eigenvalues of $(\Sigma_{i}'+sI)$ can be defined as $\lambda_{i1} + s,\lambda_{i2} + s$. This leads to the following proof:
\begin{equation}
\begin{split}
& r = k \cdot \sqrt{ \text{max(}\lambda_{i1},\lambda_{i2}\text{)} + s}, \\
& r \geq k\cdot\sqrt{s}, \\
& \pi r^2 \geq k^2\pi s,
\end{split}
\end{equation}
where $k$ is the hyperparameter that defines the confidence interval of the Gaussian. We follow the original implementation of 3DGS as $k=3$ which gives the 99.73$\%$ confidence interval. Using the value $k=3$ leads to the proof of the area of each Gaussian being at least $9\pi s$. 
\section{Toy experiments}
\label{supple:C}
\subsection{Implementation details}

In this section, we provide a detailed explanation of the 1D regression toy experiment shown in Section~\ref{subsec:sparseinit} of our main paper. We design the experiment to learn a target distribution $Y(x)$, where $x$ is in the range $[0, 10,000)$, using $N$ 1D Gaussians. Each of the $N$ 1D Gaussians includes its mean $\mu_i$ and variance $\sigma_i^2$ as learnable parameters to model a random 1D signal $Y(x)$ as the ground truth distribution. The target distribution $Y(x)$ is generated from random 10 1D Gaussian distributions. By blending $N$ 1D Gaussians with a learnable weight $w_i$, we train the parameters $\{\mu_i, \sigma_i, w_i\}$ through the L1 loss between the blended distribution from $N$ 1D Gaussians and $Y(x)$ as follows: 
\begin{equation}
    \mathcal{L} = \sum_{x} \ \left\lVert Y(x) - \sum_{i=0}^{N}  w_i \cdot \exp\left(-\frac{(x-\mu_i)^2}{2\sigma_i^2}\right) \right\rVert,
\end{equation}
For the dense-small-variance (DSV) scenario, we initialize the number of Gaussians as $N=1,000$ and randomly select $\mu_{i}$ and $\sigma_{i}$ values within the range $[0,1)$. For the dense-large-variance (DLV) scenario, we use the same number of Gaussians as $N=1,000$ but choose $\mu_{i}$ and $\sigma_{i}$ values randomly from the range $[300,301)$. For the sparse-large-variance (SLV) scenario, we set the number of initial Gaussians as $N=15$ and randomly select $\mu_{i}$ and $\sigma_{i}$ values from the range $[300,301)$. In all scenarios, we train the parameters for a total of 1,000 steps using the Adam optimizer with a learning rate of 0.01.

\begin{figure*}[t]
\begin{center}
\resizebox{1.0\linewidth}{!}
{
    \includegraphics[width=1\linewidth]{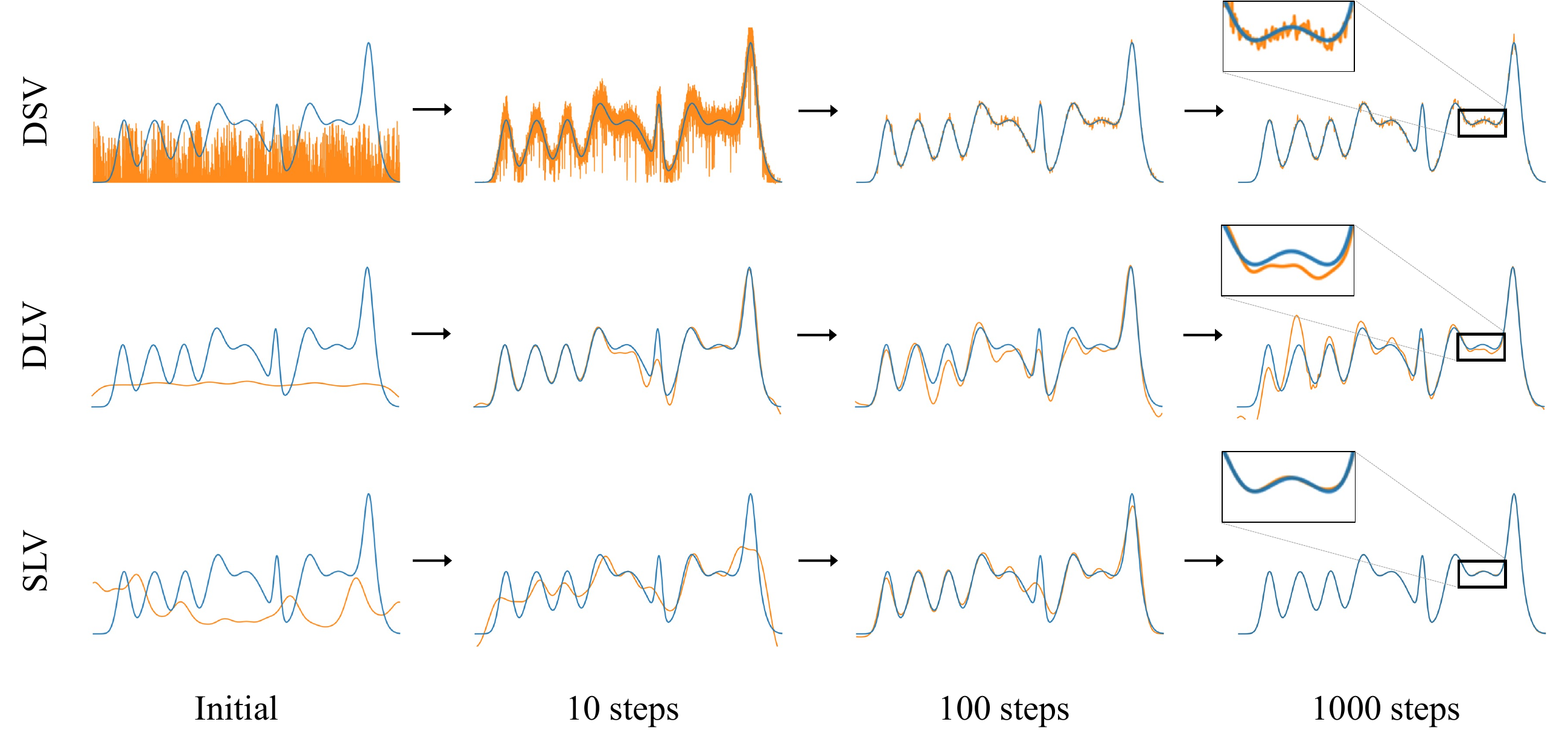}
}
\end{center}
    \caption{\textbf{Toy experiment to analyze different initialization methods.} This figure 
    visualizes the result of our toy experiment predicting the \textcolor{myblue}{target distribution} using a collection of \textcolor{myorange2}{1D Gaussians}, starting from different initialization methods. Dense-small-variance (DSV) and dense-large-variance (DLV) initialize 1,000 1D Gaussians, where DLV is initialized with large variances by adding $s$ to the initial variance. Small-large-variance (SLV) initializes 15 1D Gaussians with the same $s$ added to the initial variance. We can observe the over-fast convergence of high-frequency components on DSV initialization, which is resolved in the DLV initialization but fails to converge due to fluctuation. SLV initialization resolves both problems, learning the low-frequency components first and also converging to successfully model the \textcolor{myblue}{target distribution}.}
    \label{fig:supple_toy}
\end{figure*}

\subsection{Analysis on toy experiments}
As mentioned in Section~\ref{subsec:sparseinit}, we conduct a toy experiment in a simplified 1D regression task to examine how DSV and SLV initialization influences the optimization process. As shown in the top row of Figure~\ref{fig:supple_toy}, training with the DSV random initialization exhibits a tendency towards over-fast convergence on high frequencies. This is evident in the model's ability to capture high-frequency components after only 10 steps. However, this rapid focus on high-frequencies leads to undesired high-frequency artifacts in the final reconstruction (zoomed-in box at 1,000 steps). We hypothesize that this behavior arises from the small initial variances of the dense Gaussians, which restrict each Gaussian to model a very localized region, increasing the susceptibility to overfitting.

To verify our hypothesis, we repeat the experiment with a value $s$ added to the initial variance changing our initialization to dense-large-variance (DLV). Note that this modification effectively encourages the Gaussians to learn from wider areas. The results (middle row of Figure~\ref{fig:supple_toy}) support our hypothesis: the learned signal at 10 steps demonstrates a prioritization of low-frequency components when compared to DSV initialization. However, even with the ability to prune Gaussians by learning weights as $w_i=0$, the dense initialization causes fluctuations in the learned signal, preventing convergence after 1,000 steps. These observations highlight a crucial point: while learning from wider areas (enabled by a large variance) is necessary to prioritize low-frequency components, a dense initialization can still lead to instability and convergence issues.

To resolve both problems, we propose a sparse-large-variance (SLV) initialization. The sparsity reduces fluctuations throughout the optimization, while the large variance ensures initial focus on the low-frequency distribution. This is demonstrated in the bottom row of Figure~\ref{fig:supple_toy}, where the experiment is repeated with $N=15$ Gaussians with $s$ added to the initial variance. SLV initialization succeeds in both the prioritization of low-frequency components at 10 steps and the modeling of the true distribution with minimal errors after 1,000 steps.

Therefore, although initializing random points from the same volume which is defined as three times the size of the camera's bounding box, we initialize a significantly sparser set of 3D Gaussians. As the initial covariance of 3D Gaussian is defined based on the distances of the three nearest neighbors, sparse initialization leads to a larger initial covariance, encouraging each Gaussian to model a wider region of the scene.

\subsection{Analysis on dense-large-variance (DLV) initialization in 3DGS}
We compare the influence of different dense initialization methods through the toy experiment of learning to regress a random 1D signal with multiple 1D Gaussians. Through the toy experiment, we analyzed that while Gaussian learning from wider areas (enabled by the large initial variance) leads to prioritizing low-frequency components, the dense initialization results in instability and convergence issues. As shown in Figure~\ref{fig:supple_toy}, starting from Gaussians with large variance (dense-large-variance (DLV)) leads to prioritizing the learning of low-frequency components, compared to dense-small-variance (DSV) initialization, observed in the coarse signal in the bottom row of step 10.

Although DLV initialization succeeds in learning the low-frequency components of the true distribution first, due to the dense number of Gaussians, DLV initialization fails to converge with large fluctuations. This can be observed by comparing the learned signals from steps 10, 100, and 1000. The learned signal is already quite close to the true distribution in 10 steps but fails to converge even after a sufficient number of steps.

To verify that our findings also apply to the initialization of 3D Gaussians for 3DGS, we implement the DLV initialization and evaluate the performance of 3DGS (DLV) on the Mip-NeRF360 dataset. For this experiment, we set the initial number of Gaussians as $N=1,000,000$, equal to our setting for DSV initialization. As the initial covariance is defined by the mean distance of the three nearest neighbors, we manually scale up the initial covariance to guide the Gaussians to learn from wider areas. The results are shown in Table~\ref{exp:supp_DLV}, where the performance of DLV random initialization is between DSV and SLV random initialization, aligned with the results of our toy experiment. Note that 3DGS (SLV) in Table~\ref{exp:supp_DLV} indicates 3DGS trained using only the sparse-large-variance (SLV) random initialization of our strategy.

\begin {table*}[t]
    \caption{\textbf{Quantitative comparison on Mip-NeRF360 dataset with various initialization methods}. We compare our method with the DSV random initialization method described in the original 3DGS~\cite{kerbl20233d}, DLV random initialization, and SLV random initialization method. We report PSNR, SSIM, LPIPS. }
    \centering
    \resizebox{\textwidth}{!}{
    \begin{tabular}{l|ccc|ccc|ccc|ccc|ccc}
        \toprule
        \multirow[c]{3}{*}{ Method } & \multicolumn{15}{c}{Outdoor Scene} \\
        \cline{2-16} & \multicolumn{3}{c|}{ bicycle } & \multicolumn{3}{c|}{ flowers } & \multicolumn{3}{c|}{ garden } & \multicolumn{3}{c|}{ stump } & \multicolumn{3}{c}{ treehill }\\
        & PSNR$\uparrow$ & SSIM$\uparrow$ & LPIPS$\downarrow$ & PSNR$\uparrow$ & SSIM$\uparrow$ & LPIPS$\downarrow$ & PSNR$\uparrow$ & SSIM$\uparrow$ & LPIPS$\downarrow$ & PSNR$\uparrow$ & SSIM $\uparrow$ & LPIPS$\downarrow$ &PSNR$\uparrow$ & SSIM $\uparrow$ & LPIPS$\downarrow$  \\
        \midrule \midrule
        
        3DGS (DSV)  & 21.034  & 0.575 & 0.378  & 17.815  & 0.469 & 0.403  & 23.217  & 0.783 & 0.175 & 20.745  & 0.618 & 0.345  & 18.986 & 0.550 & 0.413\\
        
        3DGS (DLV)  & 21.595 & 0.552& 0.426 & 18.959 & 0.469 & 0.445 & 23.338 & 0.757 & 0.218 & 19.845 & 0.555 & 0.427 & 19.605 & 0.550 & 0.440  \\

        \hlrow \textbf{3DGS (SLV)} & \textbf{23.227} & \textbf{0.610} & \textbf{0.373} & \textbf{20.498} & \textbf{0.528} &\textbf{ 0.392} & \textbf{25.670} & \textbf{0.816} & \textbf{0.157} & \textbf{23.748} & \textbf{0.647} & \textbf{0.339} & \textbf{21.423} & \textbf{0.578} & \textbf{0.402} \\

        \bottomrule
        \multicolumn{16}{c}{}
        \\
        \toprule
        \multirow[c]{3}{*}{ Method } &\multicolumn{12}{c|}{Indoor Scene} & \multicolumn{3}{c}{ \multirow{2}{*}{Average}} \\
        \cline{2-13}
        & \multicolumn{3}{c|}{ room } & \multicolumn{3}{c|}{ counter } & \multicolumn{3}{c|}{ kitchen} & \multicolumn{3}{c|}{ bonsai } & \\
        
         & PSNR$\uparrow$ & SSIM$\uparrow$ & LPIPS$\downarrow$ & PSNR$\uparrow$ & SSIM$\uparrow$ & LPIPS$\downarrow$ & PSNR$\uparrow$ & SSIM$\uparrow$ & LPIPS$\downarrow$ & PSNR$\uparrow$ & SSIM $\uparrow$ & LPIPS$\downarrow$ & PSNR$\uparrow$ & SSIM $\uparrow$ & LPIPS$\downarrow$  \\
        \midrule \midrule

        3DGS (DSV) & 29.685 & \textbf{0.894} & \textbf{0.265} & 23.608 &  0.833 &  0.276  & 26.078 & 0.893 & 0.161  & 18.538 & 0.719 & 0.401 & 22.190 & 0.705 & 0.313   \\

        3DGS (DLV)  & 29.284 & 0.886 & 0.280 & 26.253 & 0.857 & 0.272 & 27.975 & 0.900 & 0.161 & 22.363 & 0.793 & 0.360 & 23.246 & 0.702 & 0.337  \\

        \hlrow 
        \textbf{3DGS (SLV)} & \textbf{29.966} & 0.892 & 0.268 & \textbf{27.473} & \textbf{0.868} & \textbf{0.260} & \textbf{29.934} & \textbf{0.915} & \textbf{0.159} & \textbf{29.132} & \textbf{0.900} & \textbf{0.251} & \textbf{25.675} & \textbf{0.750} & \textbf{0.287} \\ 
        
        \bottomrule
    \end{tabular}}

    \label{exp:supp_DLV}
    \vspace{-5pt}
\end{table*}

\section{Additional analysis}

\label{supple:D}

\subsection{Analysis on prioritized learning of low-frequency components}
In Section~\ref{subsec:motivation_SfM} of our main paper, we find that SfM initialization guides 3D Gaussians with a coarse approximation of the true distribution to robustly learn the remaining high-frequency components. In Section~\ref{subsec:sparseinit} of our main paper, we then show that sparse-large-variance (SLV) random initialization successfully guides the Gaussians to prioritize the learning of low-frequency components. To verify that our strategy successfully guides 3DGS to prioritize the learning of low-frequency components, we provide additional analysis of our strategy through the rendered images achieved from different steps of training.

\begin{figure*}[!hbt]
\begin{center}
\resizebox{1.0\linewidth}{!}
{
    \includegraphics[width=1\linewidth]{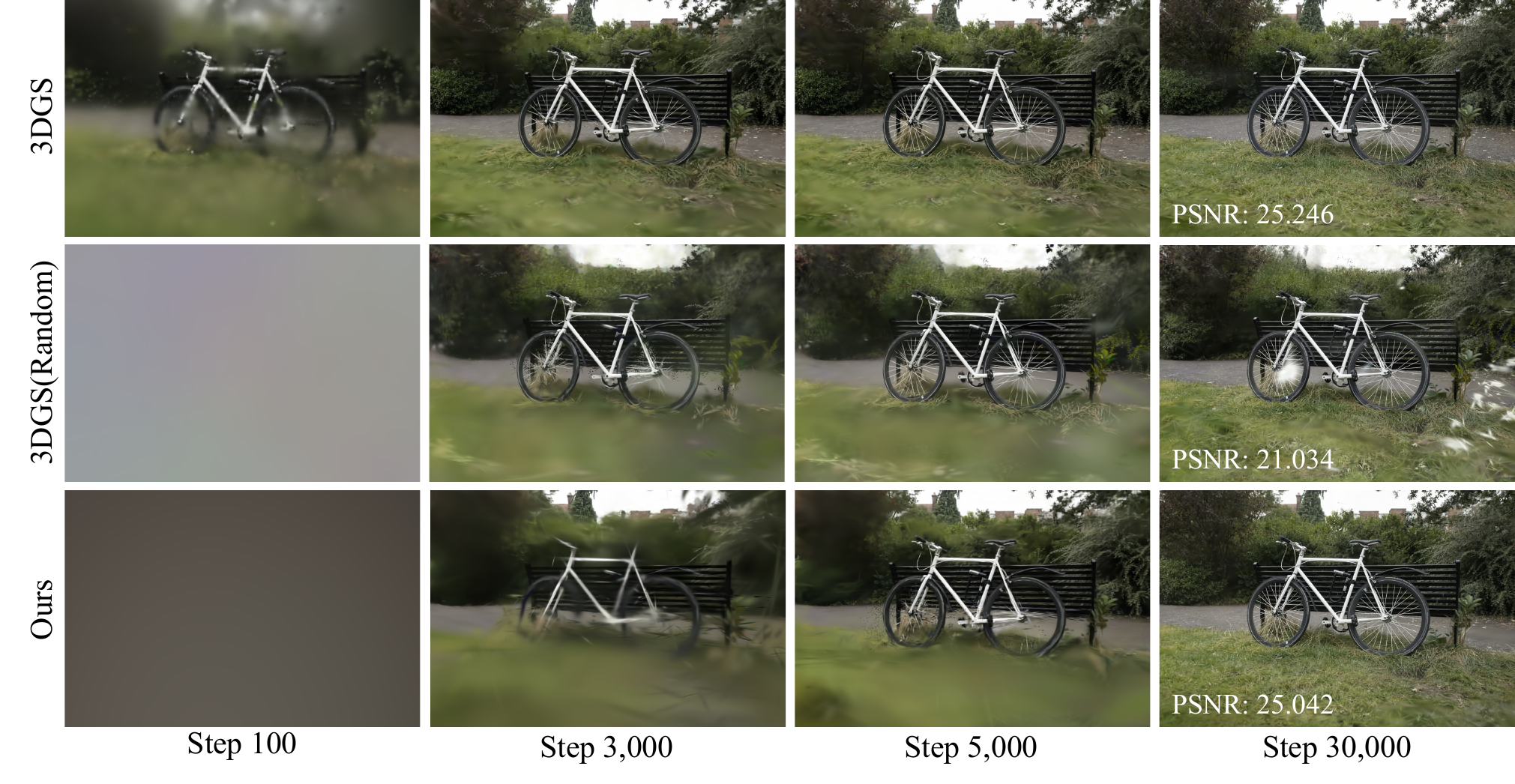}
}
\end{center}
    \caption{\textbf{Visualization of our strategy prioritizing the learning of coarse approximation.} This figure 
    visualizes the rendering result of the 'bicycle' scene from the Mip-NeRF360 dataset trained with 3DGS using different initialization methods. We show the images rendered from different steps of 100, 3,000, 5,000, and 30,000. The results of 3DGS trained from SfM initialization, DSV random initialization, and Ours are shown from top to bottom.}
    \label{fig:supple_compare_rendering}
\end{figure*}

As shown in Figure~\ref{fig:supple_compare_rendering}, we compare the rendered results of 3DGS trained with different initialization. As SfM provides a coarse approximation of the true distribution, 3DGS trained with SfM initialization (first row) directly learns the remaining high-frequency details. This tendency to directly learn the high-frequency components is highlighted in the second row where 3DGS is trained with DSV random initialization. Without any strategy to prioritize the learning of low-frequency components, the high-frequency components such as the edges of the bench and bicycle are already learned as shown in the rendered image of step 3,000. In contrast, our strategy (last row) successfully guides the Gaussians to first model the scene's coarse approximation. At step 3,000, high-frequency details are absent, resulting in smoothed appearances for the grass, road, and background trees.

\begin{figure*}[t]
\begin{center}
\resizebox{1.0\linewidth}{!}
{
    \includegraphics[width=1\linewidth]{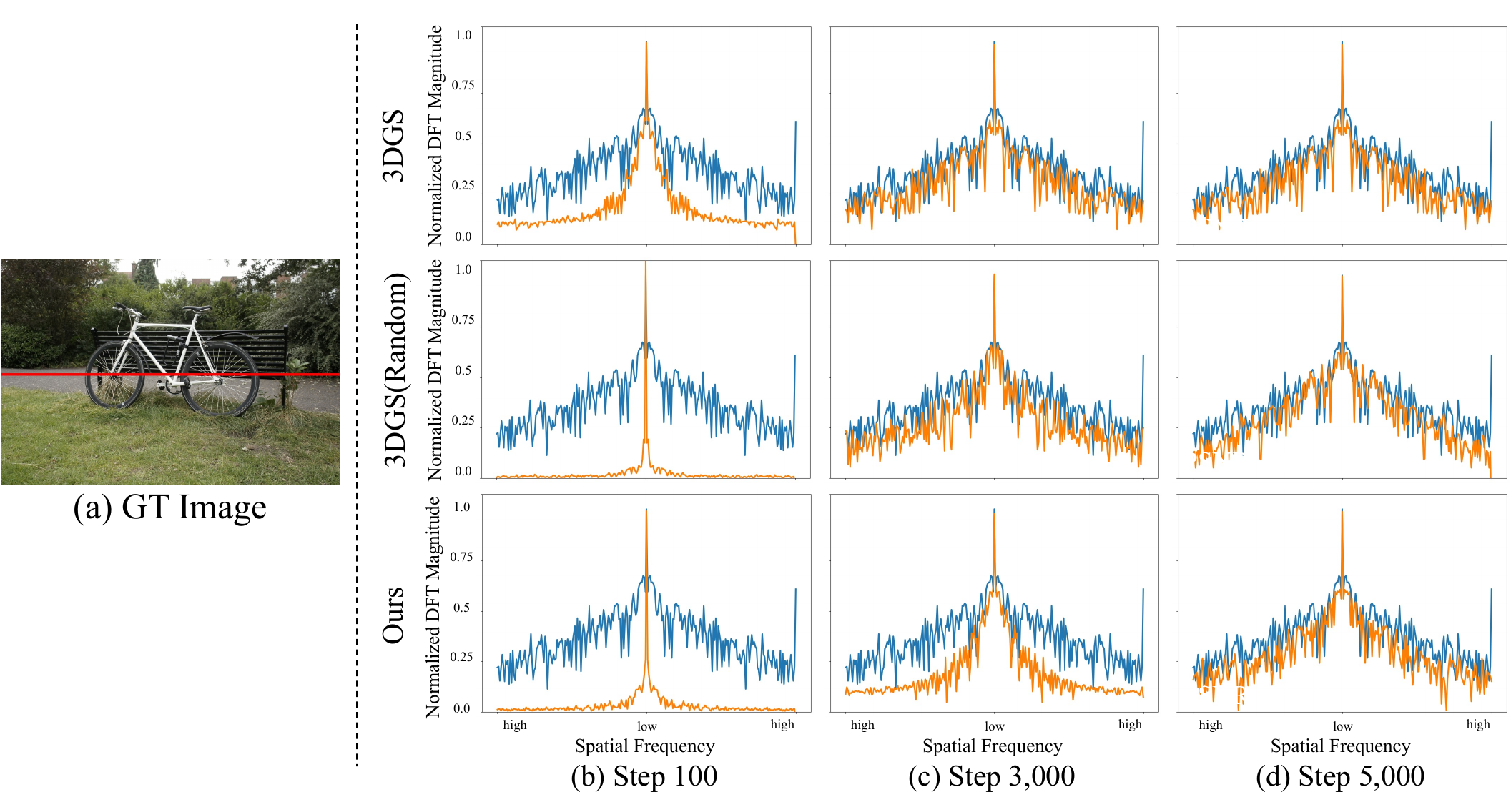}
}
\end{center}
    \caption{\textbf{Images from Figure~\ref{fig:supple_compare_rendering} analyzed in the frequency domain.}}
    \label{fig:supple_fft}
\end{figure*}

In Figure~\ref{fig:supple_fft}, we further analyze the rendered images of Figure~\ref{fig:supple_compare_rendering} in the frequency domain. We sample a \textcolor{myred}{horizontal line} from the images as in (a) and perform FFT~\cite{nussbaumer1982fast}. From (b) to (d), the normalized DFT magnitude of the transformed signal from the \textcolor{myorange2}{GT image} and the \textcolor{myblue}{rendered image} is shown in sequential timesteps of 100, 3,000, and 5,000.

As mentioned in Section~\ref{subsec:motivation_SfM} of our main paper, 3DGS (top row) starts from a coarse approximation of the signal, which can be observed by the low-frequency components already being captured in step 100 ((b), top row). This is more evident when compared to the transformed signal of 3DGS (DSV) and Ours at step 100 ((b), middle and bottom row), where the low-frequency components are largely different from the GT signal. Aligned with our analysis of the toy experiment in Section~\ref{subsec:sparseinit} of our main paper, DSV random initialization exhibits a tendency towards over-fast convergence on high-frequencies. This can be observed by comparing (b) and (c), where the high-frequency components are quickly learned in (c). This leads to high-frequency artifacts after optimization, as shown in the rendered image (step 30,000, middle row) in Figure~\ref{fig:supple_compare_rendering}.

On the other hand, our strategy successfully guides the Gaussians to prioritize the learning of low-frequency components. When comparing (b), (c), and (d) of the last row, the low-frequency components are learned in (c) and then the high-frequency details are captured in (d). Note that this prioritization of learning low-frequency components makes the overall training process similar to starting with SfM initialization. The similarity is further highlighted by comparing the transformed signals of 3DGS in (b) and Ours in (c).

\subsection{Ablation study on warm up phase}
\label{supple:ablation}
\paragraph{Ablation settings for low-pass filter control.}
\label{supple:lowpass}
To demonstrate the effectiveness of our progressive Gaussian low-pass filter control strategy, we employ three different decreasing functions of convex, linear, and concave to control the Gaussian low-pass filter value $s$. 
Different from our strategy, where the value $s$ is defined adaptively by image height, width, and the number of Gaussians $N$ at each time step, the remaining functions are manually defined to achieve $s=300$ at step 0 and $s=0.3$ at about 3,000 steps across all scenes. The intuition behind this design is based on our analysis that our adaptive Gaussian low-pass filter value reaches $0.3$ between 2,000-3,000 steps. Also, we empirically find that the initial Gaussian low-pass filter value $s > 300$ offers no significant improvement, only making the overall computation inefficient. Based on these findings, we define the max value of the Gaussian low-pass filter as $s=300$.

For the convex function, we use the following formula for $s$ scheduling:
\begin{equation}
   s = \max(7^{-\frac{x}{1000}}*300, 0.3).
\end{equation}
For the linear function, we use the following formula for $s$ scheduling:
\begin{equation}
   s = \max(300-0.0997084x, 0.3).
\end{equation}
For the concave function, we use the following formula for $s$ scheduling:
\begin{equation}
   s = \max(300*(1+7^{-3}-7^{\frac{x-3000}{1000}}), 0.3).
\end{equation}

\begin{figure*}[h]
\parbox{.48\linewidth}{
\begin{center}
\includegraphics[width=1\linewidth]{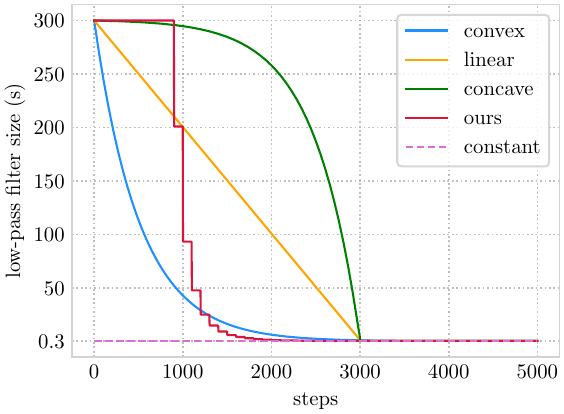}
\end{center}
    \caption{\textbf{Illustration of different Gaussian low-pass filter value formulas.} }
    \label{fig:lowpassablation}
}
\hfill
\parbox{.48\linewidth}{
\begin{center}
\includegraphics[width=1\linewidth]{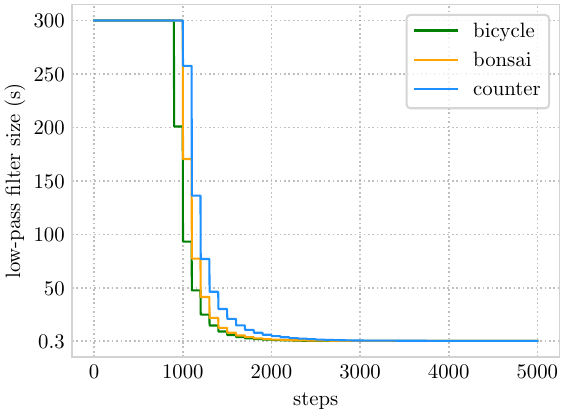}
\end{center}
    \caption{\textbf{Illustration of our Gaussian low-pass filter value formula.} }
    \label{fig:lowpassours}
    
}
\end{figure*}

The illustration of different Gaussian low-pass value formulas is shown in Figure~\ref{fig:lowpassablation} and Figure~\ref{fig:lowpassours} where our formula is adaptively defined, showing different functions for each scene.

\paragraph{Comparison.}
In Table~\ref{fig:lowpassablation}, we show a more detailed ablation study on the components belonging to the warm up phase. Additional experiments regarding (1) fixing learning rate during the warm up phase (2) the other progressive low-pass filter control functions mentioned above are shown here. Results show that our choice of dynamically adjusting the low-pass filter value based on the number of Gaussians achieves the best performance compared to the others. Also as can be seen in the Table, each of fixing learning rate, SLV initialization and \oursplit \ solely contributes to the performance improvement, verifying the effect of our proposed methods. Compared to the original DSV random initialization, we achieve an significant average PSNR improvement over 5dB.

\begin{table}[!hbt]
  \caption{\textbf{Ablation study on warm up phase.} 
  }
  \label{tab:c2f_abl}
  \centering
  \begin{tabular}{c|c|c|c|ccc}
\toprule
 \multirow{2}{*}{Low-pass filter}& \multirow{2}{*}{Init.} & \multirow{2}{*}{Fixed learning rate} &\multirow{2}{*}{\oursplit} &\multicolumn{3}{c}{Mip-NeRF360} \\
 & & & &PSNR$\uparrow$ & SSIM$\uparrow$ & LPIPS$\downarrow$\\
 \midrule\midrule
Constant &  DSV & \ding{55} & \ding{55} & 22.190	& 0.704 & 0.313 \\
Constant &  SLV & \ding{55} & \ding{55} & 25.675 & 0.749 & 0.288 \\
Progressive (Convex) & SLV& \ding{55} &\ding{55}  &25.799 & 0.756 & 0.286\\
Progressive (Linear)& SLV &\ding{55}  &\ding{55}  &25.898 & 0.754 & 0.290\\
Progressive (Concave) & SLV &\ding{55}  &\ding{55}  &25.974 & 0.754 & 0.288\\
\midrule
Ours & SLV &\ding{55}  &\ding{55}  & 26.180 & 0.761 & 0.280\\
Ours & SLV &\ding{51}  &\ding{55}  &26.362& 0.791& 0.241\\
\hlrow \textbf{Ours} & \textbf{SLV}& \ding{51} & \ding{51}&  \textbf{27.228} & \textbf{0.807} & \textbf{0.229}\\
\bottomrule
  \end{tabular}
\end{table}

\begin {table*}[!hbt]
    \caption{\textbf{Quantitative comparison on Mip-NeRF360 dataset in sparse SfM points setting}. We compare our method with 3DGS in sparse SfM points setting, We report PSNR, SSIM and LPIPS.
    }
    \centering
    \resizebox{\textwidth}{!}{
    \begin{tabular}{l|c|ccc|ccc|ccc|ccc|ccc}
        \toprule \multirow[c]{3}{*}{ Method } & \multirow[c]{3}{*}{ \shortstack{SfM \\ points}} & \multicolumn{15}{c}{Outdoor Scene} \\
        \cline{3-17}
         & & \multicolumn{3}{c|}{ bicycle } & \multicolumn{3}{c|}{ flowers } & \multicolumn{3}{c|}{ garden } & \multicolumn{3}{c|}{ stump } & \multicolumn{3}{c}{ treehill }\\
        & & PSNR$\uparrow$ & SSIM$\uparrow$ & LPIPS$\downarrow$ & PSNR$\uparrow$ & SSIM$\uparrow$ & LPIPS$\downarrow$ & PSNR$\uparrow$ & SSIM$\uparrow$ & LPIPS$\downarrow$ & PSNR$\uparrow$ & SSIM $\uparrow$ & LPIPS$\downarrow$ &PSNR$\uparrow$ & SSIM $\uparrow$ & LPIPS$\downarrow$  \\
        \midrule \midrule

        3DGS & cluster  & 24.542& 0.703& 0.289& 21.044& 0.569& 0.365& 26.883& 0.852&0.118& 25.774& 0.735& 0.250& 22.102& 0.601& 0.376\\

        3DGS & <10\%  & 24.456& 0.700& 0.294& 21.260& 0.580& 0.360& 26.832& 0.850&0.127& 26.165& 0.744& 0.247& 22.260& 0.612& 0.365\\

        3DGS & full  &\textbf{ {25.246}}& \textbf{{0.771}} & \textbf{{ 0.205}}&  21.520&   0.605&  0.336& \textbf{{27.410}}& \textbf{{0.868}}&\textbf{{ 0.103}}&  26.550& \underline{0.775}&  {0.210}&  22.490& \textbf{{0.638}}& {\textbf{0.317}}\\
        \midrule
        
        \hlrow Ours & cluster  & 25.211& 0.767& 0.211& \underline{21.704}& \underline{0.617} & \underline{0.321}& \underline{27.166}& \underline{0.862}& \textbf{0.103}& \underline{26.816}& \underline{0.775}& \underline{0.204}& \underline{22.589}& 0.630& 0.328\\

        \hlrow Ours & <10\%  & \underline{25.214} & \underline{0.769} & \underline{0.208} &  \textbf{{21.858}}& \textbf{{0.619}}& \textbf{{0.320}}& {27.130} & {0.861} &  \underline{0.107}& \textbf{{26.899}}& \textbf{{0.777}}& \textbf{{0.202}}& {\textbf{22.618}}& \underline{0.632}& \underline{0.327}\\

        \bottomrule
        \multicolumn{17}{c}{}
        \\
        \toprule
        \multirow[c]{3}{*}{ Method } & \multirow[c]{3}{*}{ \shortstack{SfM \\ points}} & \multicolumn{12}{c|}{Indoor Scene} & \multicolumn{3}{c}{ \multirow{2}{*}{Average}} \\
        \cline{3-14} & & \multicolumn{3}{c|}{ room } & \multicolumn{3}{c|}{ counter } & \multicolumn{3}{c|}{ kitchen} & \multicolumn{3}{c|}{ bonsai }  \\
        
        & & PSNR$\uparrow$ & SSIM$\uparrow$ & LPIPS$\downarrow$ & PSNR$\uparrow$ & SSIM$\uparrow$ & LPIPS$\downarrow$ & PSNR$\uparrow$ & SSIM$\uparrow$ & LPIPS$\downarrow$ & PSNR$\uparrow$ & SSIM $\uparrow$ & LPIPS$\downarrow$ & PSNR$\uparrow$ & SSIM $\uparrow$ & LPIPS$\downarrow$  \\
        \midrule \midrule

        3DGS & cluster  & 29.919& 0.899& 0.255& 28.256& 0.893& 0.227& 30.127& 0.917&0.140& 31.247& 0.932& 0.220& 26.655& 0.789& 0.249\\

        3DGS & <10\%  &  {30.795} & 0.908& 0.243&  28.523&  0.896& 0.223& 30.120& 0.915&0.140& 31.625& 0.934& 0.221& 26.893& 0.793& 0.247\\

        3DGS  & full  &  \underline{30.632}&  {\textbf{0.914}}&  \textbf{{0.220}}&  \underline{28.700}& {\textbf{ 0.905}}&  \textbf{{0.204}}&   30.317& \underline{ {0.922}}&  \textbf{{0.129}}&  \textbf{{31.980}}& \textbf{ {0.938}}&  \textbf{{0.205}}& {27.205} &  \textbf{{0.815}} & \textbf{{0.214}}  \\
        \midrule

        \hlrow Ours & cluster  & {30.363}& 0.909& 0.233& 28.594& 0.901& 0.211& \underline{31.136}& \textbf{0.923}& \underline{0.133}& 31.695& \underline{0.935}& 0.217& \underline{27.253}& \underline{0.813}& 0.218\\

        \hlrow Ours & <10\% & {\textbf{30.843}}& \underline{0.912}& \underline{0.230}&\textbf{ {28.703}} & \underline{0.902}& \underline{0.209}&\textbf{{31.364}}& \underline{{0.922}}& \underline{0.133} &  \underline{31.829}& \textbf{{0.938}}& \underline{0.214}&  \textbf{{27.384}}& \textbf{{0.815}}& \underline{0.217}\\

        \bottomrule
    \end{tabular}}

    \label{tab:sparse_mip360}
\end{table*}

\subsection{Ablation on sparse SfM initialization}
\label{supple:sparse_sfm}
\
In Table~\ref{tab:sparse_mip360}, we compare two methods for creating sparse SfM points to reduce the uncertainty of the SfM point cloud and use our coarse-to-fine method with the SfM point cloud: cluster and <10\%. For the cluster, we use HDBSCAN~\cite{mcinnes2017accelerated} to cluster the points based on the location of the SfM point cloud, and use the centroids   of the clusters for the sparse SfM initialization. We conduct experiments using the default hyperparameters in scikit-learn~\cite{kramer2016scikit} for HDBSCAN~\cite{mcinnes2017accelerated}. For <10\%, we create sparse SfM points by using only the top 10\% of points utilizing the reprojection error value that comes out with the SfM output. We confirm that our method achieves additional performance improvements by using the SfM points together.

\begin {table*}[h]
    \caption{\textbf{Quantitative comparison on Mip-NeRF360 dataset in noisy initial SfM point cloud settings}. We compare 3DGS method with different noisy inital SfM point cloud. We report PSNR, SSIM, LPIPS. }
    \centering
    \resizebox{\textwidth}{!}{
    \begin{tabular}{l|ccc|ccc|ccc|ccc|ccc}
        \toprule
        \multirow[c]{3}{*}{Initialization} &\multicolumn{15}{c}{Outdoor Scene} \\
        \cline{2-16} & \multicolumn{3}{c|}{ bicycle } & \multicolumn{3}{c|}{ flowers } & \multicolumn{3}{c|}{ garden } & \multicolumn{3}{c|}{ stump } & \multicolumn{3}{c}{ treehill }\\
        & PSNR$\uparrow$ & SSIM$\uparrow$ & LPIPS$\downarrow$ & PSNR$\uparrow$ & SSIM$\uparrow$ & LPIPS$\downarrow$ & PSNR$\uparrow$ & SSIM$\uparrow$ & LPIPS$\downarrow$ & PSNR$\uparrow$ & SSIM $\uparrow$ & LPIPS$\downarrow$ &PSNR$\uparrow$ & SSIM $\uparrow$ & LPIPS$\downarrow$  \\
        \midrule \midrule
        
        SfM& 25.246& 0.771& 0.205& 21.520& 0.605& 0.336& 27.410& 0.868& 0.103& 26.550& 0.775& 0.210& 22.490& 0.638& 0.317\\
        
        SfM + ${\epsilon}$& 24.289& 0.692& 0.307& 21.028& 0.565& 0.374& 26.515& 0.846& 0.133& 26.028& 0.743& 0.252& 21.728& 0.607& 0.380\\

        SfM+constant& 23.619& 0.625& 0.358& 21.139& 0.569&0.364& 25.663& 0.809& 0.163& 23.382& 0.641& 0.335& 21.989& 0.593& 0.380\\

        \bottomrule
        \multicolumn{16}{c}{}
        \\
        \toprule
        \multirow[c]{3}{*}{Initialization} & \multicolumn{12}{c|}{Indoor Scene} &  \multicolumn{3}{c}{ \multirow{2}{*}{Average}}  \\
        \cline{2-13} & \multicolumn{3}{c|}{ room } & \multicolumn{3}{c|}{ counter } & \multicolumn{3}{c|}{ kitchen} & \multicolumn{3}{c|}{ bonsai } &  \\
        
        & PSNR$\uparrow$ & SSIM$\uparrow$ & LPIPS$\downarrow$ & PSNR$\uparrow$ & SSIM$\uparrow$ & LPIPS$\downarrow$ & PSNR$\uparrow$ & SSIM$\uparrow$ & LPIPS$\downarrow$ & PSNR$\uparrow$ & SSIM $\uparrow$ & LPIPS$\downarrow$ & PSNR$\uparrow$ & SSIM $\uparrow$ & LPIPS$\downarrow$  \\
        \midrule \midrule

        SfM& 30.632& 0.914& 0.220& 28.700&  0.905&  0.204& 30.317& 0.922& 0.129& 31.980& 0.938& 0.205& 27.205& 0.815& 0.214\\

        SfM + ${\epsilon}$& 30.740& 0.901& 0.260& 23.872& 0.838& 0.285& 24.477& 0.877& 0.175& 25.823& 0.875& 0.265& 24.944& 0.771& 0.270\\

        SfM+constant& 30.790& 0.899& 0.264& 28.041& 0.875& 0.250& 30.203& 0.914& 0.145& 31.006& 0.924& 0.230& 26.204& 0.761& 0.277\\ 
        
        \bottomrule
    \end{tabular}}

    \label{exp:supp_noiseSfM}
    \vspace{-5pt}
\end{table*}
\subsection{Analysis on the impact of noise on the initial SfM point cloud}
\begin{figure*}[!hbt]
\begin{center}
\resizebox{0.7\linewidth}{!}
{
    \includegraphics[width=1\linewidth]{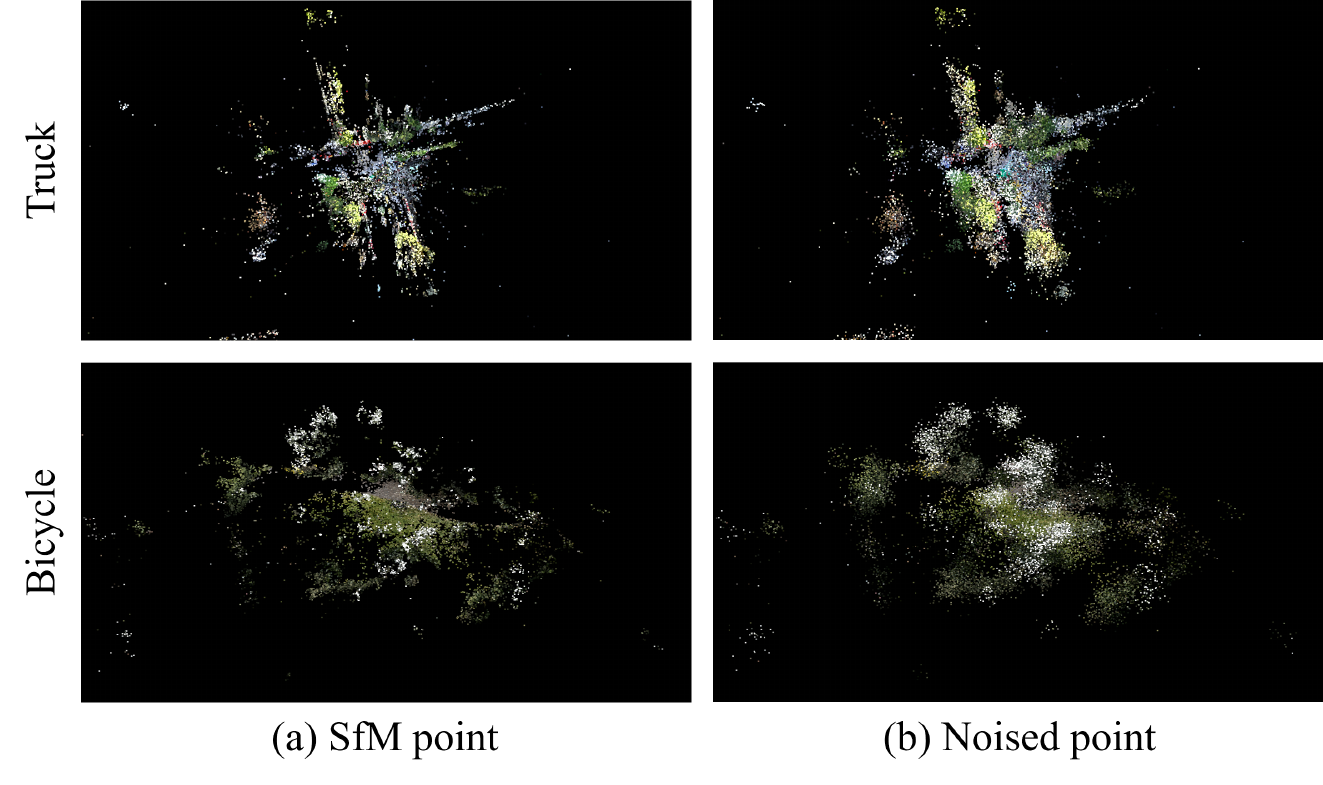}
}
\end{center}
    \caption{\textbf{Visualization of SfM points and points with noise $\epsilon$.}}
    \label{fig:supple_noise_vis}
\end{figure*}

In Table~\ref{tab:noise_sfm}, we present detailed results to further investigate the ability of the 3DGS optimization scheme to transport Gaussians to the correct 3DGS locations on Mip-NeRF360 datasets. Here, we conduct this experiments by adding random noise $\epsilon \sim \mathcal{N}(0, 1)$ and constant systematic noise, whose value equals 2, to the initial SfM points. The results shown in Table~\ref{exp:supp_noiseSfM} prove that 3DGS strongly depends on the initial point. Figure~\ref{fig:supple_noise_vis} shows initial SfM points and points with noise $\epsilon$. Even with the small amount of noise, 3DGS fails to move to the correct position.

\begin{figure*}[!hbt]
\begin{center}
\resizebox{1.0\linewidth}{!}
{
    \includegraphics[width=1\linewidth]{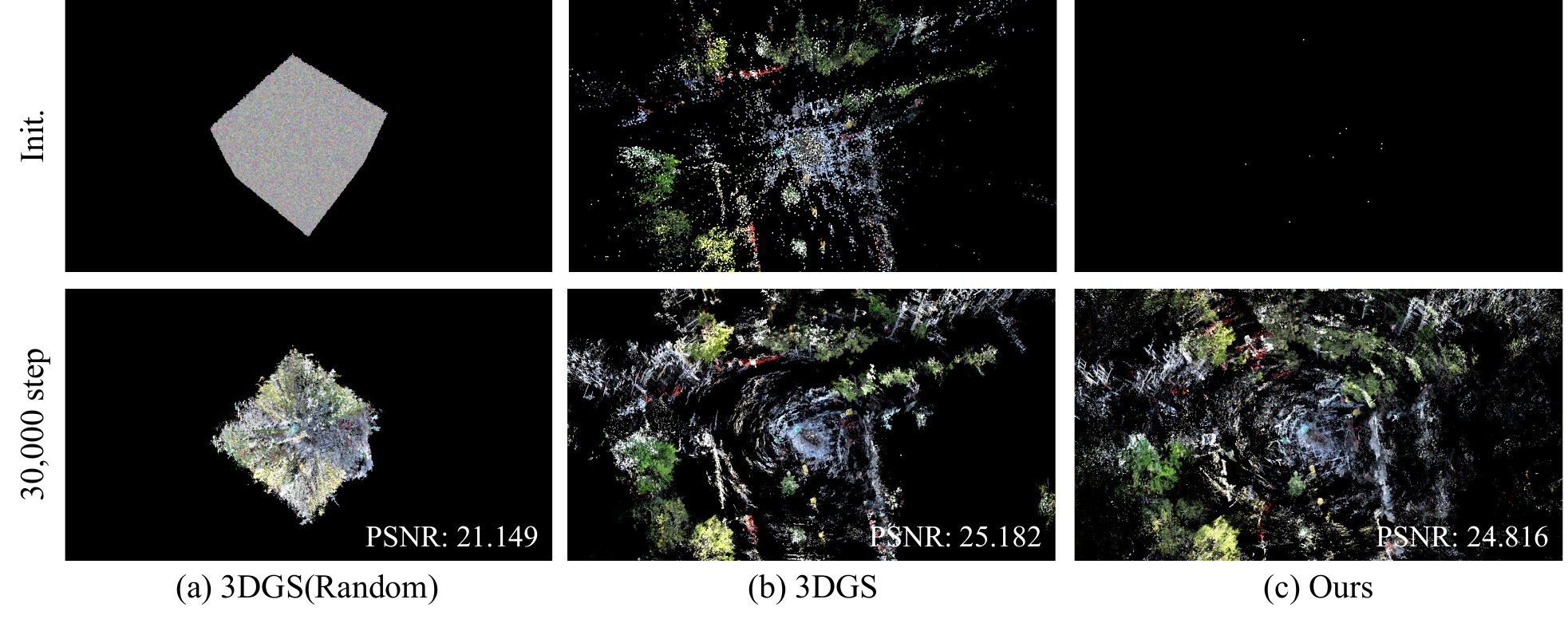}
}
\end{center}
    \caption{\textbf{Displacements of Gaussians from initial positions}}
    \label{fig:supple_movement}
\end{figure*}

\subsection{Analysis on movement of each Gaussian}
\label{subsec:gaussian_distance}
To observe the movement of the each Gaussian, we measure how far the Gaussian moves from its initial position during the training. An additional parameter is incorporated to record the initial position, ensuring that even when Gaussians are split or cloned, the initial position parameters are retained. Then, the movement is calculated as difference between the final position of each Gaussian after training and its respective initial position. 

We conduct analysis on ``Truck'' scene of Tanks\&Temples dataset, comparing the settings of 3DGS with SfM point initialization, 3DGS with random initialization, and our method. The mean, standard deviation, and the top 1\% values of the movements is shown in Table~\ref{tab:movement}. Additionally, Figure~\ref{fig:supple_movement} shows the overall scene from the same camera viewpoint for each experiment to observe the differences in distribution of overall Gaussians between the beginning and 30,000 step. In case such as 3DGS with SfM initialization and random initialization, the positions of the Gaussians does not change significantly. However, our method shows substantial changes in comparison.

\subsection{Expanding Scene bound}
\label{supp:expanding_bound}

As discussed in Section~\ref{subsec:motivation_3DGS} and can be seen in supplementary materials Section~\ref{subsec:gaussian_distance}, Gaussians struggle to move further from the initialized locations. Therefore, in Table~\ref{tab:supp_expand_scene}, we compare the results according to the initial scene bound and our method using the \oursplit \ algorithm on the Mip-NeRF360 dataset to see if it is sufficient to initialize with a sufficiently wide scene bound. Extent $\times$ 3 and Extent $\times$ 10 are the initial scene bounds created by multiplying the camera extent given in the dataset by 3 and 10, respectively, and 3 is used in the original 3DGS. As can be seen in Extent x 10, it is confirmed that simply starting with a wide initial scene bound does not yield sufficient performance.

\subsection{Detailed results of Oracle scene bound experiments}

To find out if the performance degradation of using noisy or random point cloud \textit{\textbf{solely}} comes from losing the information of where the Gaussians should be located, we conduct a simple experiment of training 3DGS with random point cloud sampled from the oracle scene bounds. Specifically, here we assume the oracle scene bounds as the final scene bound defined by the final positions of the Gaussians, which are trained from point cloud achieved from SfM. Table~\ref{exp:supp_oracle} presents the results for each scene. The experiment reveals that SfM initialization provides additional benefits to 3DGS beyond merely placing the Gaussians in proper locations, as starting from random point cloud even in oracle scene bounds leads to sub-optimal results compared to the one starting from SfM-initialized point cloud.

\begin {table*}[!hbt]
    \caption{\textbf{Quantitative comparison on Mip-NeRF360 dataset within different scene bound settings}. We compare our method (SLV) with 3DGS using random initialization within different scale bounding box. We report PSNR, SSIM, LPIPS. }
    \centering
    \resizebox{\textwidth}{!}{
    \begin{tabular}{l|ccc|ccc|ccc|ccc|ccc}
        \toprule
        \multirow[c]{3}{*}{Scene bound} & \multicolumn{15}{c}{Outdoor Scene} \\
        \cline{2-16} & \multicolumn{3}{c|}{ bicycle } & \multicolumn{3}{c|}{ flowers } & \multicolumn{3}{c|}{ garden } & \multicolumn{3}{c|}{ stump } & \multicolumn{3}{c}{ treehill }\\
        & PSNR$\uparrow$ & SSIM$\uparrow$ & LPIPS$\downarrow$ & PSNR$\uparrow$ & SSIM$\uparrow$ & LPIPS$\downarrow$ & PSNR$\uparrow$ & SSIM$\uparrow$ & LPIPS$\downarrow$ & PSNR$\uparrow$ & SSIM $\uparrow$ & LPIPS$\downarrow$ &PSNR$\uparrow$ & SSIM $\uparrow$ & LPIPS$\downarrow$  \\
        \midrule \midrule
        
        Extent$\times$3& 21.034& 0.575& 0.378& 17.815& 0.469& 0.403& 23.217& 0.783& 0.175& 20.745& 0.618& 0.345& 18.986& 0.550& 0.413\\
        
        Extent$\times$10& 23.864& 0.643& 0.354& 20.974& 0.553& 0.381& 25.862& 0.823& 0.151& 24.876& 0.696& 0.300& 22.067& 0.590& 0.394\\

        \hlrow Ours& \textbf{25.042}& \textbf{0.747}& \textbf{0.238}& \textbf{21.762}& \textbf{0.616}&\textbf{0.324}&\textbf{ 26.884}& \textbf{0.854}& \textbf{0.114}& \textbf{26.680}& \textbf{0.768}& \textbf{0.215}& \textbf{22.528}& \textbf{0.621}& \textbf{0.342}\\

        \bottomrule
        \multicolumn{16}{c}{}
        \\
        \toprule
        \multirow[c]{3}{*}{Scene bound} &\multicolumn{12}{c|}{Indoor Scene}&  \multicolumn{3}{c}{ \multirow{2}{*}{Average}}\\
        \cline{2-13} & \multicolumn{3}{c|}{ room } & \multicolumn{3}{c|}{ counter } & \multicolumn{3}{c|}{ kitchen} & \multicolumn{3}{c|}{ bonsai }  \\
        
        & PSNR$\uparrow$ & SSIM$\uparrow$ & LPIPS$\downarrow$ & PSNR$\uparrow$ & SSIM$\uparrow$ & LPIPS$\downarrow$ & PSNR$\uparrow$ & SSIM$\uparrow$ & LPIPS$\downarrow$ & PSNR$\uparrow$ & SSIM $\uparrow$ & LPIPS$\downarrow$ & PSNR$\uparrow$ & SSIM $\uparrow$ & LPIPS$\downarrow$  \\
        \midrule \midrule

        Extent$\times$3& 29.685& 0.894& 0.265& 23.608&  0.833&  0.276& 26.078& 0.893& 0.161& 18.538& 0.719& 0.401& 22.190& 0.704& 0.313\\

        Extent$\times$10& 29.834& 0.889& 0.276& 27.441& 0.863& 0.267& 30.424& 0.913& 0.147& 30.482& 0.917& 0.242& 26.203& 0.765& 0.279\\

        \hlrow Ours& \textbf{30.809}& \textbf{0.906}& \textbf{0.247}& \textbf{28.529}& \textbf{0.895}& \textbf{0.223}& \textbf{31.270}& \textbf{0.920}& \textbf{0.137}& \textbf{31.547}& \textbf{0.934}& \textbf{0.218}& \textbf{27.228}& \textbf{0.807}& \textbf{0.229}\\ 
        
        \bottomrule
    \end{tabular}}

    \label{tab:supp_expand_scene}
    \vspace{-5pt}
\end{table*}
\begin {table*}[h]
    \caption{\textbf{Quantitative comparison on Mip-NeRF360 dataset in oracle scene bound settings}. We compare 3DGS method with random initialization in paper setting, SfM initialization and random initialization in oracle scene bound. We report PSNR, SSIM, LPIPS. }
    \centering
    \resizebox{\textwidth}{!}{
    \begin{tabular}{l|ccc|ccc|ccc|ccc|ccc}
        \toprule
        \multirow[c]{3}{*}{Initialization}&\multicolumn{15}{c}{Outdoor Scene} \\
        \cline{2-16} & \multicolumn{3}{c|}{ bicycle } & \multicolumn{3}{c|}{ flowers } & \multicolumn{3}{c|}{ garden } & \multicolumn{3}{c|}{ stump } & \multicolumn{3}{c}{ treehill }\\
        & PSNR$\uparrow$ & SSIM$\uparrow$ & LPIPS$\downarrow$ & PSNR$\uparrow$ & SSIM$\uparrow$ & LPIPS$\downarrow$ & PSNR$\uparrow$ & SSIM$\uparrow$ & LPIPS$\downarrow$ & PSNR$\uparrow$ & SSIM $\uparrow$ & LPIPS$\downarrow$ &PSNR$\uparrow$ & SSIM $\uparrow$ & LPIPS$\downarrow$  \\
        \midrule \midrule
        
        3DGS (Random)& 21.034& 0.575& 0.378& 17.815& 0.469& 0.403& 23.217& 0.783& 0.175& 20.745& 0.618& 0.345& 18.986& 0.550& 0.413\\
        
        3DGS (SfM)& 25.246& 0.771& 0.205& 21.520& 0.605& 0.336& 27.410& 0.868& 0.103& 26.550& 0.775& 0.210& 22.490& 0.638& 0.317\\

        3DGS (Oracle)& 24.227& 0.664& 0.328& 21.353& 0.582&0.356& 26.791& 0.843& 0.131& 24.634& 0.678& 0.318& 22.219& 0.600& 0.378\\

        \bottomrule
        \multicolumn{16}{c}{}
        \\
        \toprule
        \multirow[c]{3}{*}{Initialization} &\multicolumn{12}{c|}{Indoor Scene}&  \multicolumn{3}{c}{ \multirow{2}{*}{Average}} \\
        \cline{2-13} & \multicolumn{3}{c|}{ room } & \multicolumn{3}{c|}{ counter } & \multicolumn{3}{c|}{ kitchen} & \multicolumn{3}{c|}{ bonsai } & \\
        
         & PSNR$\uparrow$ & SSIM$\uparrow$ & LPIPS$\downarrow$ & PSNR$\uparrow$ & SSIM$\uparrow$ & LPIPS$\downarrow$ & PSNR$\uparrow$ & SSIM$\uparrow$ & LPIPS$\downarrow$ & PSNR$\uparrow$ & SSIM $\uparrow$ & LPIPS$\downarrow$ & PSNR$\uparrow$ & SSIM $\uparrow$ & LPIPS$\downarrow$  \\
        \midrule \midrule

        3DGS (Random)& 29.685& 0.894& 0.265& 23.608&  0.833&  0.276& 26.078& 0.893& 0.161& 18.538& 0.719& 0.401& 22.190& 0.704& 0.313\\

        3DGS (SfM)& 30.632& 0.914& 0.220& 28.700& 0.905& 0.204& 30.317& 0.922& 0.129& 31.980& 0.938& 0.205& 27.205& 0.815& 0.214\\

        3DGS (Oracle)& 30.192& 0.895& 0.265& 27.812& 0.873& 0.251& 30.797& 0.917& 0.140& 30.303& 0.916& 0.238& 26.481& 0.774& 0.267\\ 
        
        \bottomrule
    \end{tabular}}

    \label{exp:supp_oracle}
    \vspace{-5pt}
\end{table*}

\subsection{Comparison of computational resources}
\label{supple:computation}
In this section, we now compare the computational resources of Plenoxels~\cite{yu2021plenoxels}, InstantNGP~\cite{muller2022instant}, 3DGS~\cite{kerbl20233d}, 3DGS with random initialization, and 3DGS with our strategy. We evaluate different methods on the Mip-NeRF360~\cite{barron2022mip}, Tanks\&Temples~\cite{knapitsch2017tanks}, and Deep Blending~\cite{hedman2018deep} dataset.

\vspace{-10pt}
\paragraph{Training time.} Note that when pre-defined camera poses are available, methods other than 3DGS~\cite{kerbl20233d} can bypass the time-consuming Structure-from-Motion (SfM) process. To highlight this overhead, in Table~\ref{exp:train_time} we compare the training time of 3DGS with the SfM process taken into account. For 3DGS, we find that the number of Gaussians in each training step has a direct correlation to the overall training time - more Gaussians lead to longer training time.
\begin {table}[!hbt]
    \caption{\textbf{Training time.}\tablefootnote{Except for the SfM processing time and training times of 3DGS(Random) and Ours, we copied the values from 3DGS~\cite{kerbl20233d}. Note that 3DGS utilized a single A6000 GPU and we utilized a single RTX 3090 GPU for evaluation.}}
    
    \centering
    \resizebox{0.7\textwidth}{!}{
    \begin{tabular}{c|c|c|c}
        \toprule & Mip-NeRF360 &  Tanks\&Temples & Deep Blending \\
        \midrule
        
        Plenoxels  & 25m49s & 25m5s & 27m49s \\

        INGP-Base  & 5m37s & 5m26s & 6m31s  \\

        INGP-Big & 7m30s & 6m59s & 8m \\

        3DGS\tablefootnote{The processing time of SfM is denoted with a $\dagger$.} &   41m33s $+$ (19m9s){$^\dagger$}& 26m54s $+$ (19m43s){$^\dagger$}&   36m2s $+$ (6m23s){$^\dagger$} \\

        3DGS(Random) & 28m22s & 18m31s & 29m33s \\
        \hlrow Ours & 31m34s & 15m13s & 27m51s \\
        \bottomrule
    \end{tabular}
    }

    \label{exp:train_time}
\end{table}

\clearpage
\subsection{Limitations}
\label{supp:limit}
\begin{figure*}[!hbt]
\begin{center}
\resizebox{1.0\linewidth}{!}
{
    \includegraphics[width=1\linewidth]{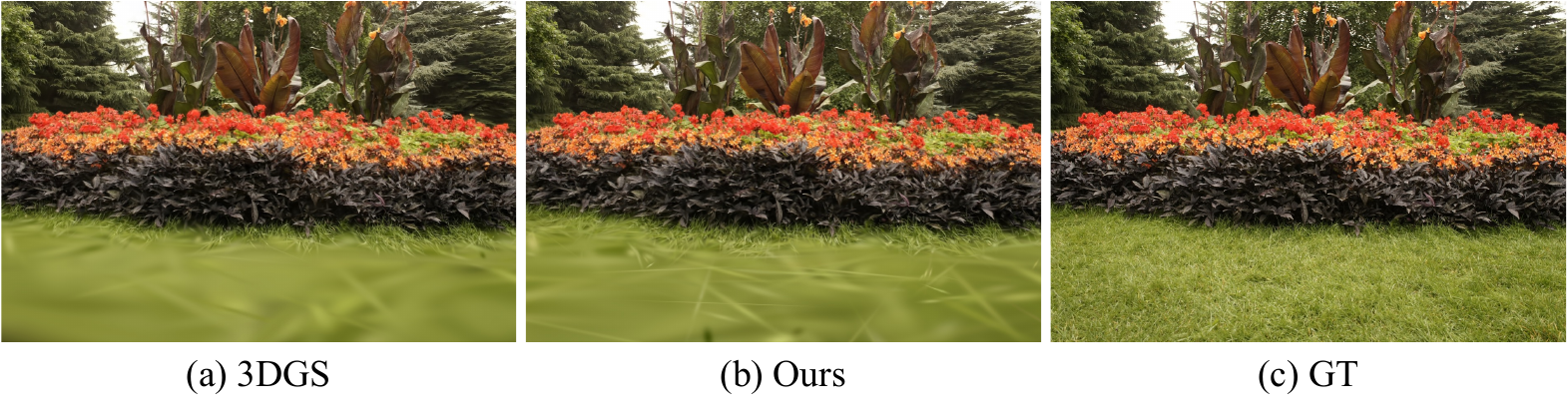}
}
\end{center}
    \caption{\textbf{Visualization of failure cases.} This figure shows the rendering results of the ``flowers'' scene of the Mip-NeRF360 dataset from Ours and 3DGS. When compared to the GT image shown in (c), the front of the images of (a) and (b) lack high-frequency details.}
    \label{fig:limitation}
\end{figure*}

Although \textbf{RAIN-GS} robustly guides 3D Gaussians to model the scene even from randomly initialized point cloud, our novel strategy is not without limitations. As our strategy prioritizes learning the coarse approximation, it sometimes fails to detect the need for further densification to capture high-frequency details. This happens especially for regions where the L1 rendering loss cannot distinguish between a coarse approximation and a high-frequency true distribution. An example of this limitation is shown in Figure~\ref{fig:limitation}, where the grass region in the front of the rendered image lacks high-frequency details compared to the ground truth image. However, we find that this limitation also appears when training 3DGS with SfM initialized point cloud, which is also shown to start optimization from a coarse approximation of the true distribution in our main paper. We posit that this limitation mainly comes from the lack of supervision, using the L1 rendering loss as the main signal. Although adding additional supervision from depth or error maps could open up new possibilities to resolve these limitations, we leave this direction as future work.

\vspace{-10pt}

\clearpage
\section{Additional qualitative results}
\label{supple:E}
\begin{figure}[!hbt]
\begin{center}
{
    \includegraphics[width=1\linewidth]{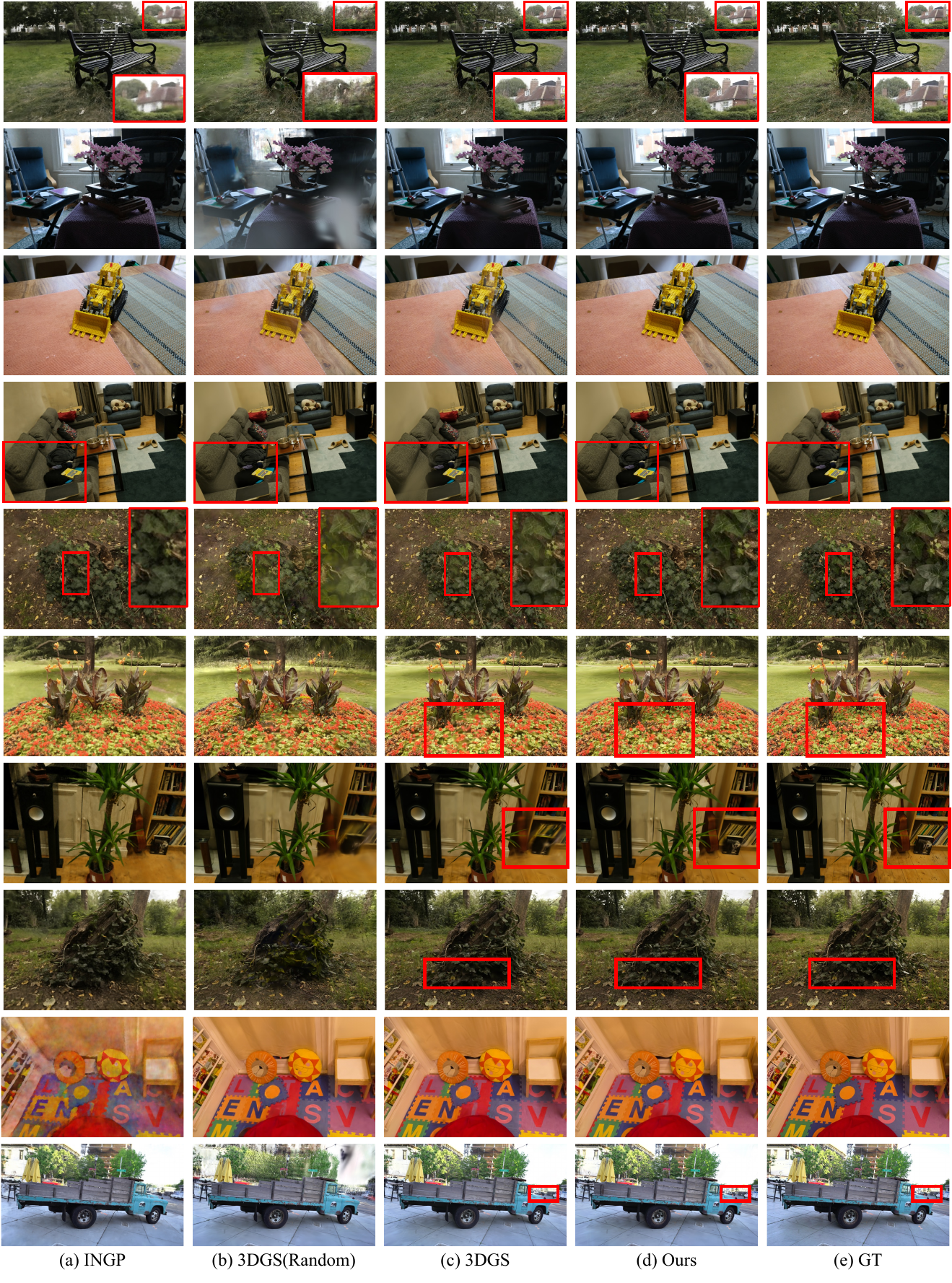}
}
\end{center}
\caption{\textbf{Additional qualitative results.}}
\label{fig:supple_qual}
\end{figure}

We show additional qualitative results in Figure~\ref{fig:supple_qual}.

\clearpage

\begin{figure}[!hbt]
\begin{center}
{
    \includegraphics[width=1\linewidth]{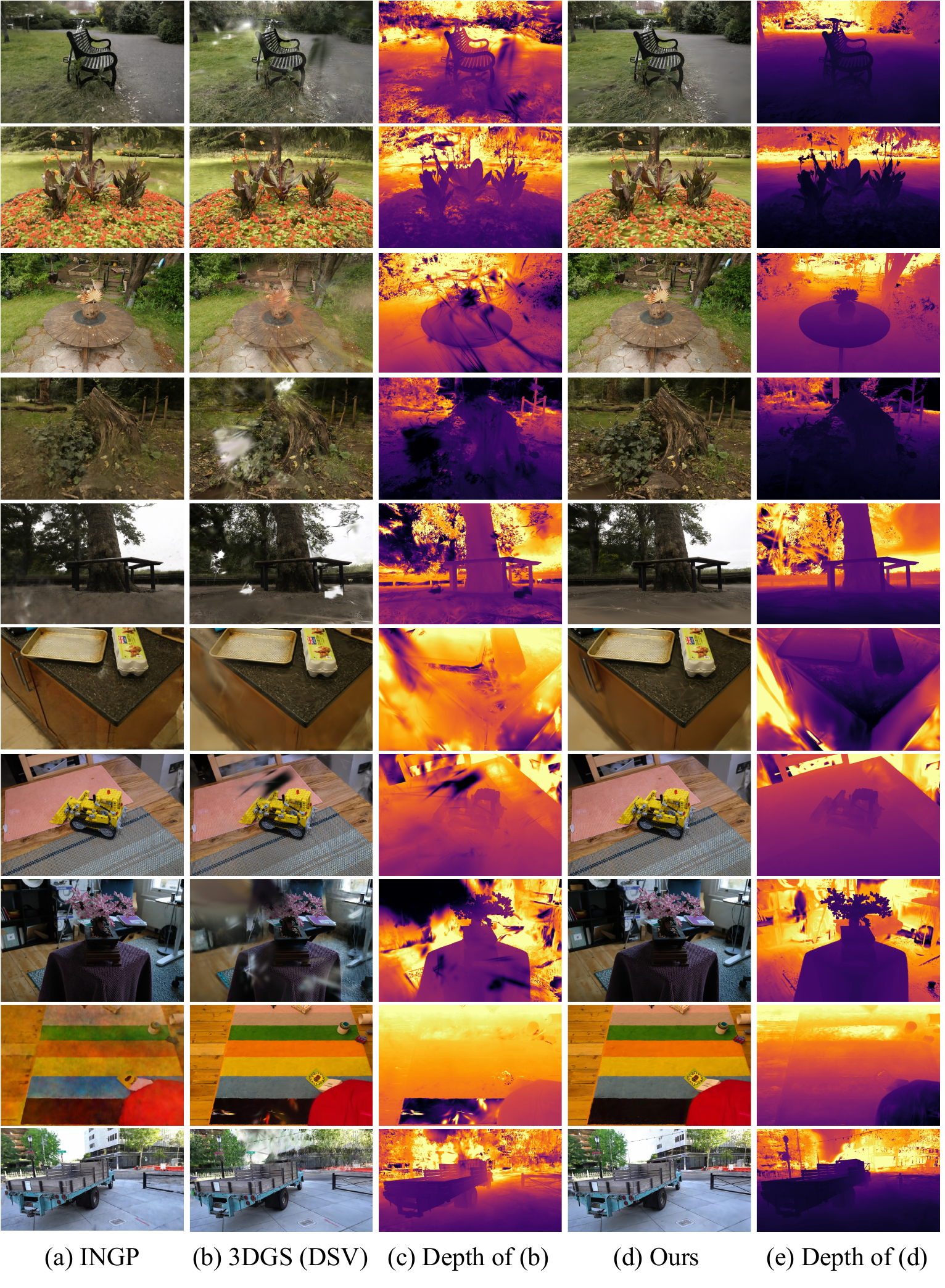}
}
\end{center}
\caption{\textbf{Additional qualitative results with depth.}}
\label{fig:supple_qual2}
\end{figure}
\clearpage
\section{Additional results of 3DGS in sparse view settings}
\label{supple:F}

\begin{figure}[!hbt]
\begin{center}
{
    \includegraphics[width=1\linewidth]{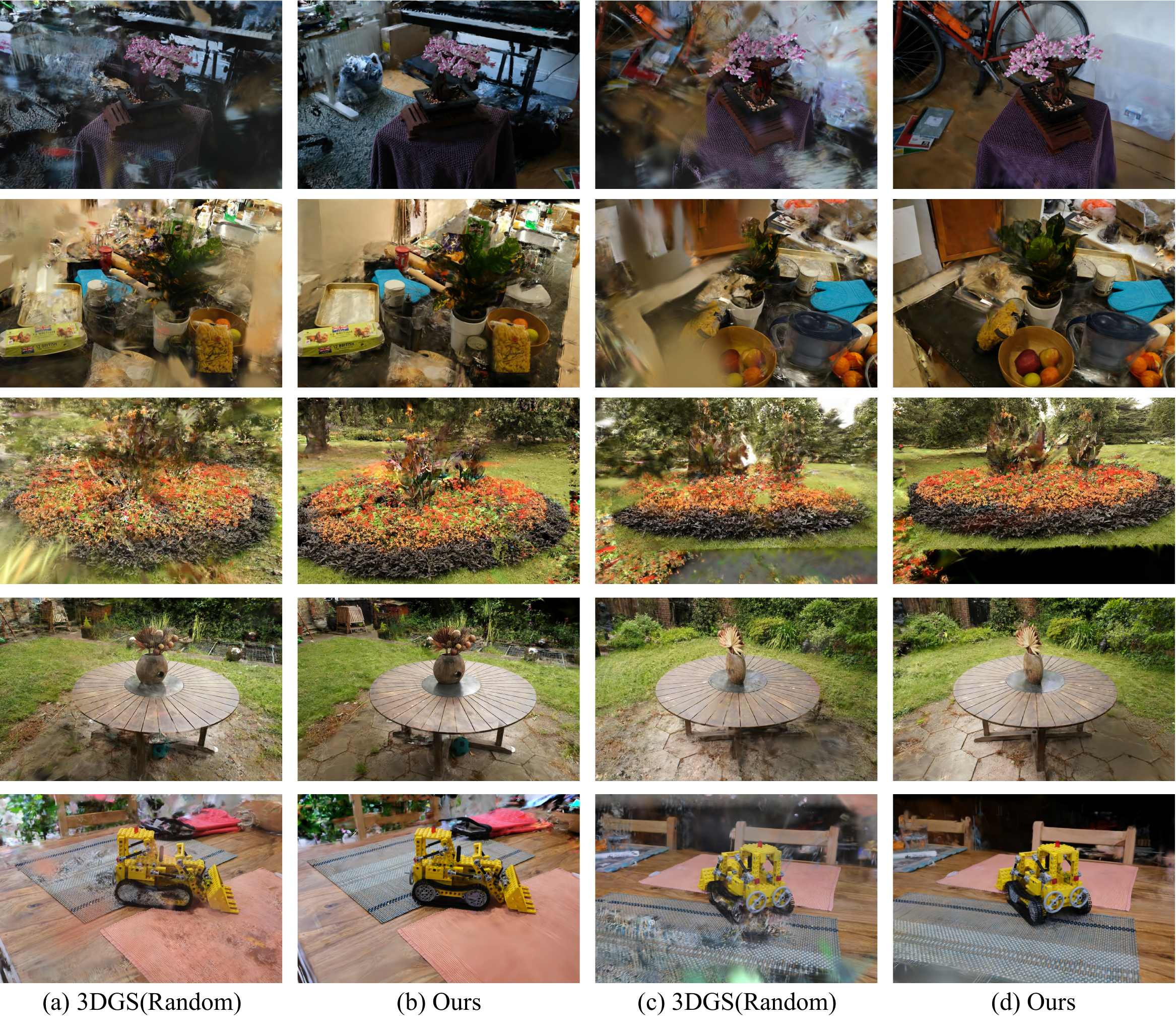}
}
\end{center}
\caption{\textbf{Qualitative results in sparse view settings.}}
\label{fig:supple_qual_sparse}
\end{figure}

Due to the impressive performance shown by NeRF, various follow-ups have been proposed, specifically to succeed in learning NeRF with a limited number of images~\cite{niemeyer2021regnerf, yang2023freenerf, jung2023selfevolving, song2024darf, kwak2023geconerf}. However, due to the requirement of accurate pose and initial point cloud of 3DGS, the task of successfully guiding 3D Gaussians with a limited number of images becomes a much more complicated task. In this section, we evaluate our strategy in training 3DGS with few-shot images, where SfM struggles to converge. In Table~\ref{exp:supp_fewshot} and Figure~\ref{fig:supple_qual_sparse}, we show quantitative and qualitative results comparing 3DGS(Random) and Ours in the sparse view setting in the Mip-NeRF360 dataset~\cite{barron2022mip}. To show the effectiveness of our strategy when SfM struggles to converge, after excluding every 8th image for evaluation, we train the model on randomly sampled 10$\%$ of the remaining images. Ours outperforms 3DGS trained with random initialization both qualitatively and quantitatively, demonstrating the potential of our work to be extended to the task of training 3DGS with few-shot images. In Figure~\ref{fig:supple_qual_sparse}, it is shown that the rendered results of Ours much more robustly models the real geometry of the scene.

\begin {table*}[h]
    \caption{\textbf{Quantitative comparison on Mip-NeRF360 dataset in sparse-view setting}. We compare our method with the random initialization method described in the original 3DGS~\cite{kerbl20233d}. We report PSNR, SSIM, LPIPS.}
    \centering
    \resizebox{\textwidth}{!}{
    \begin{tabular}{l|ccc|ccc|ccc|ccc|ccc}
        \toprule
        \multirow[c]{3}{*}{ Method } & \multicolumn{15}{c}{Outdoor Scene} \\
        \cline{2-16}  & \multicolumn{3}{c|}{ bicycle } & \multicolumn{3}{c|}{ flowers } & \multicolumn{3}{c|}{ garden } & \multicolumn{3}{c|}{ stump } & \multicolumn{3}{c}{ treehill }\\
        & PSNR$\uparrow$ & SSIM$\uparrow$ & LPIPS$\downarrow$ & PSNR$\uparrow$ & SSIM$\uparrow$ & LPIPS$\downarrow$ & PSNR$\uparrow$ & SSIM$\uparrow$ & LPIPS$\downarrow$ & PSNR$\uparrow$ & SSIM $\uparrow$ & LPIPS$\downarrow$ &PSNR$\uparrow$ & SSIM $\uparrow$ & LPIPS$\downarrow$  \\
        \midrule \midrule
        
        3DGS(Random)  & 11.787   & \textbf{0.170}  & 0.637  &  10.948  & 0.118  & \textbf{0.648} & 16.752 & 0.377 & 0.444 & \textbf{15.239} & \textbf{0.194}  & \textbf{0.602} & 12.675 & \textbf{0.206} & 0.599  \\

        \hlrow Ours & \textbf{12.436}& 0.157& \textbf{0.636}&  \textbf{11.546}& \textbf{0.120}& 0.649& \textbf{17.955}& \textbf{0.463}& \textbf{0.425}& 14.675& 0.142& 0.614& \textbf{12.820}& 0.191& \textbf{0.597}\\

        \bottomrule
        \multicolumn{16}{c}{}
        \\
        \toprule
        
        \multirow[c]{3}{*}{ Method } & \multicolumn{12}{c|}{Indoor Scene}&  \multicolumn{3}{c}{ \multirow{2}{*}{Average}} \\
        \cline{2-13} & \multicolumn{3}{c|}{ room } & \multicolumn{3}{c|}{ counter } & \multicolumn{3}{c|}{ kitchen} & \multicolumn{3}{c|}{ bonsai } &  \\
        
         & PSNR$\uparrow$ & SSIM$\uparrow$ & LPIPS$\downarrow$ & PSNR$\uparrow$ & SSIM$\uparrow$ & LPIPS$\downarrow$ & PSNR$\uparrow$ & SSIM$\uparrow$ & LPIPS$\downarrow$ & PSNR$\uparrow$ & SSIM $\uparrow$ & LPIPS$\downarrow$ & PSNR$\uparrow$ & SSIM $\uparrow$ & LPIPS$\downarrow$  \\
        \midrule \midrule

        3DGS(Random)  & 18.864 & 0.702 & \textbf{0.418} & 15.311 & 0.575 & 0.475 & 18.181 & 0.629 & 0.384 & 13.246 & 0.508 & 0.538 & 14.778 & 0.386 & 0.527 \\

        \hlrow Ours & \textbf{19.804}& \textbf{0.722}& 0.424& \textbf{18.201}& \textbf{0.637}& \textbf{0.451}& \textbf{21.474}& \textbf{0.718}& \textbf{0.359}& \textbf{20.664}& \textbf{0.721}& \textbf{0.408}& \textbf{16.620}& \textbf{0.430}& \textbf{0.507}\\
        \bottomrule
        
    \end{tabular}}

    \label{exp:supp_fewshot}
    \vspace{-5pt}
\end{table*}

\clearpage
\medskip
{\small
\bibliographystyle{unsrt}
\bibliography{egbib}
}

\end{document}